\documentclass[10pt,twocolumn,letterpaper]{article}

\usepackage[pagenumbers]{cvpr} 

\usepackage{graphicx}
\usepackage{amsmath}
\usepackage{amssymb}
\usepackage{bm}
\usepackage{nicefrac}
\usepackage{booktabs}
\usepackage{multirow}
\usepackage{makecell}
\usepackage{array}
\usepackage{enumitem}
\usepackage{gensymb}  
\usepackage{microtype}  
\usepackage[percent]{overpic} 
\usepackage{rotating}
\usepackage{adjustbox}
\usepackage[binary-units]{siunitx}
\usepackage{pifont}
\usepackage[dvipsnames]{xcolor}
\usepackage[pagebackref,breaklinks,colorlinks]{hyperref}
\usepackage{xr}  

\hypersetup{
    citecolor=ForestGreen,
    urlcolor=MidnightBlue,
}

\usepackage[outline]{contour}
\contourlength{0.6pt} 

\newcommand{\PAR}[1]{\vskip4pt \noindent{\bf #1~}}
\renewcommand{\*}[1]{\bm{\mathrm{#1}}}

\renewcommand{\b}[1]{\textbf{#1}}

\newcommand{\0}{\phantom{0}}

\newcommand{\xmark}{\ding{55}}%
\newcommand{\cmark}{\ding{51}}%
\newcommand{\cbullet}[1]{{\large $\color{#1}\bullet$}}

\newcommand{\supp}{Appendix} 
\newif\ifproceedings  
\newif\ifwithsupp  
\newif\ifsupponly
\proceedingsfalse
\withsupptrue
\supponlyfalse

\DeclareSIUnit\pixel{pixel}

\DeclareFontFamily{U}{mathb}{}
\DeclareFontShape{U}{mathb}{m}{n}{
  <-5.5> mathb5
  <5.5-6.5> mathb6
  <6.5-7.5> mathb7
  <7.5-8.5> mathb8
  <8.5-9.5> mathb9
  <9.5-11.5> mathb10
  <11.5-> mathb12
}{}
\DeclareSymbolFont{mathb}{U}{mathb}{m}{n}
\DeclareMathSymbol{\drsh}{3}{mathb}{"EB}

\usepackage[capitalize]{cleveref}
\crefname{section}{Sec.}{Secs.}
\Crefname{section}{Section}{Sections}
\Crefname{table}{Table}{Tables}
\crefname{table}{Tab.}{Tabs.}

\def\cvprPaperID{4255} 
\def\confName{CVPR}
\def\confYear{2023}

\begin{document}

\setlength{\abovedisplayskip}{7.0pt plus 1.0pt minus 5.0pt}
\setlength{\belowdisplayskip}{7.0pt plus 1.0pt minus 5.0pt}
\setlength{\belowdisplayshortskip}{6.0pt plus 1.0pt minus 3.0pt}

\title{OrienterNet: Visual Localization in 2D Public Maps with Neural Matching}

\author{%
Paul-Edouard Sarlin$^{1}$\hspace{.07in}%
Daniel DeTone$^{2}$\hspace{.07in}%
Tsun-Yi Yang$^{2}$\hspace{.07in}%
Armen Avetisyan$^{2}$\hspace{.07in}%
Julian Straub$^{2}$\\
Tomasz Malisiewicz$^{2}$\hspace{.045in}%
Samuel Rota Bulo$^{2}$\hspace{.045in}%
Richard Newcombe$^{2}$\hspace{.045in}%
Peter Kontschieder$^{2}$\hspace{.045in}%
Vasileios Balntas$^{2}$%
\vspace{0.05in}\\
$^{1}$ ETH Zurich\hspace{0.2in}
$^{2}$ Meta Reality Labs
}

\ifsupponly\else 
\maketitle

\begin{abstract}
Humans can orient themselves in their 3D environments using simple 2D maps.
Differently, algorithms for visual localization mostly rely on complex 3D point clouds that are expensive to build, store, and maintain over time.
We bridge this gap by introducing OrienterNet, the first deep neural network that can localize an image with sub-meter accuracy using the same 2D semantic maps that humans use.
OrienterNet estimates the location and orientation of a query image by matching a neural Bird's-Eye View with open and globally available maps from OpenStreetMap, enabling anyone to localize anywhere such maps are available.
OrienterNet is supervised only by camera poses but learns to perform semantic matching with a wide range of map elements in an end-to-end manner.
To enable this, we introduce a large crowd-sourced dataset of images captured across 12 cities from the diverse viewpoints of cars, bikes, and pedestrians.
OrienterNet generalizes to new datasets and pushes the state of the art
in both robotics and AR scenarios.
The code and trained model will be released publicly.
\end{abstract}

\section{Introduction}

As humans, we intuitively understand the relationship between what we see and what is shown on a map of the scene we are in. 
When lost in an unknown area, we can accurately pinpoint our location by carefully comparing the map with our surroundings using distinct geographic features. 

Yet, algorithms for accurate visual localization are typically complex, as they rely on image 
matching and require detailed 3D point clouds and visual descriptors~\cite{lowe2004distinctive,rublee2011orb,superpoint,irschara2009structure,sattler2012improving,Lynen2020IJRR,sarlin2019coarse}.
Building 3D maps with LiDAR or photogrammetry~\cite{snavely2006photo,frahm2010building,agarwal2011building,schoenberger2016sfm,moulon2016openmvg}
is expensive at world scale
and requires costly, freshly-updated data to capture temporal changes in visual appearance.
3D maps are also expensive to store, as they are orders of magnitude larger than basic 2D maps.
This prevents executing localization on-device and usually requires costly cloud infrastructure.
Spatial localization is thus a serious bottleneck for the large-scale deployment of robotics and augmented reality devices.
This disconnect between the localization paradigms of humans and machines
leads to the important research question of
\emph{How can we teach machines to localize from basic 2D maps like humans do?}

\begin{figure}[t]
    \centering
    \includegraphics[width=\linewidth]{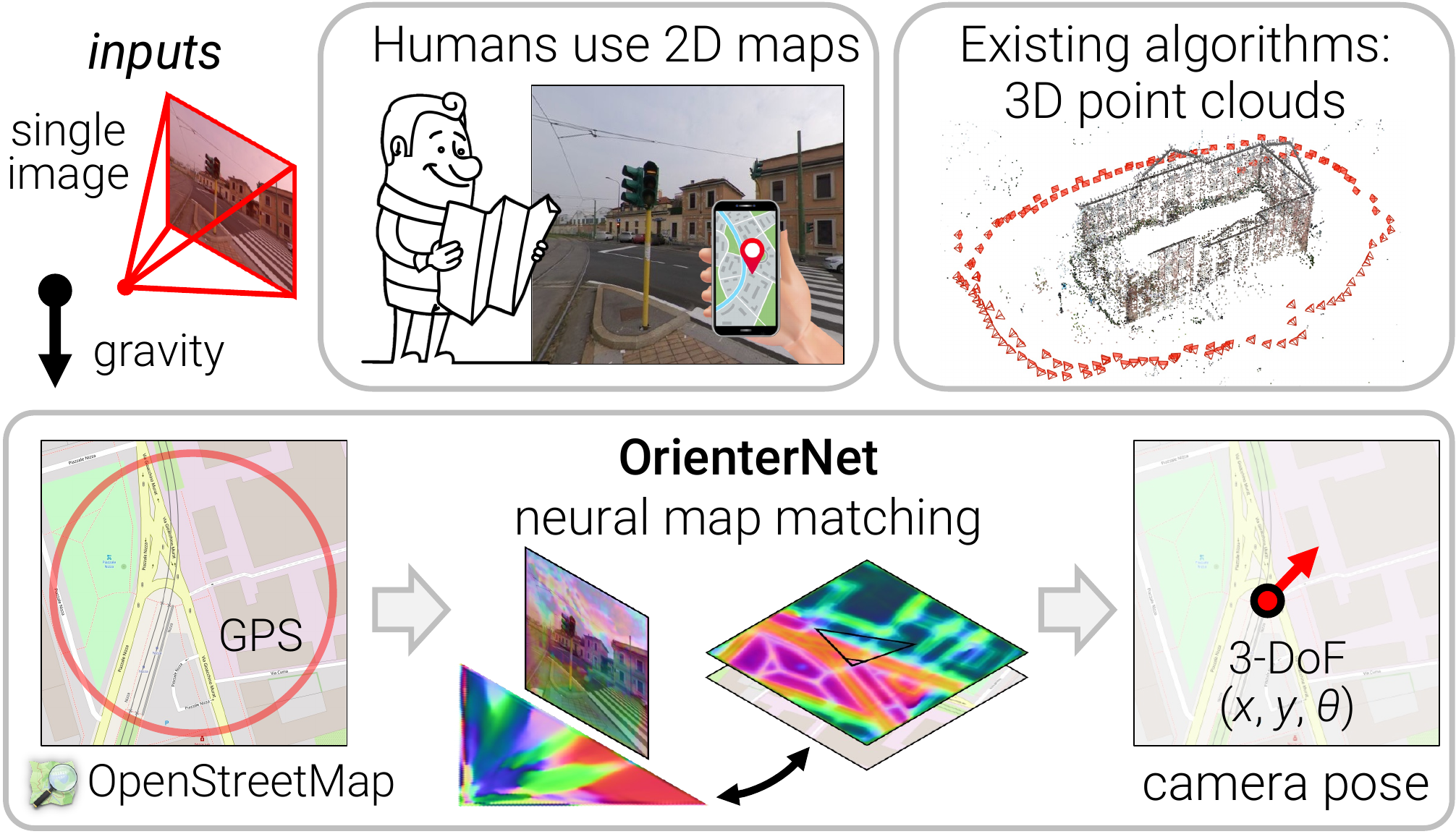}
    \caption{\textbf{Towards human-like localization.}
    Humans can easily orient themselves with basic 2D maps while state-of-the-art algorithms for visual localization require complex 3D cues.
    OrienterNet can localize an image using only compact maps from OpenStreetMap by matching Bird's-Eye View and neural maps.
    }%
    \label{fig:teaser}%
\end{figure}

This paper introduces the first approach that can localize single images and image sequences with sub-meter accuracy given the same maps that humans use. 
These \emph{planimetric} maps encode only the location and coarse 2D shape of few important objects but not their appearance nor height.
Such maps are extremely compact, up to ${10}^4$ times smaller in size than 3D maps, and can thus be stored on mobile devices and used for on-device localization within large areas. 
We demonstrate these capabilities with OpenStreetMap~(OSM)~\cite{OpenStreetMap}, an openly accessible and community-maintained world map, enabling anyone to localize anywhere for free.
This solution does not require building and maintaining costly 3D maps over time nor collecting potentially sensitive mapping data.

Concretely, our algorithm estimates the 3-DoF pose, as position and heading, of a calibrated image in a 2D map. 
The estimate is probabilistic and can therefore be fused with an inaccurate GPS prior or across multiple views from a multi-camera rig or image sequences. 
The resulting solution is significantly more accurate than consumer-grade GPS sensors and reaches accuracy levels closer to the traditional pipelines based on feature matching~\cite{sattler2012improving,sarlin2019coarse}.

Our approach, called OrienterNet, is a deep neural network that 
mimics the way humans orient themselves in their environment when looking at maps, \ie, by matching the metric 2D map with a mental map derived from visual observations~\cite{O'Keefe1978,lobben2007navigational}. 
OrienterNet learns to compare visual and semantic data in an end-to-end manner, supervised by camera poses only. 
This yields accurate pose estimates by leveraging the high diversity of semantic classes exposed by OSM, from roads and buildings to objects like benches and trash cans. 
OrienterNet is also fast and highly interpretable. 
We train a single model that generalizes well to previously-unseen cities and across images taken by various cameras from diverse viewpoints -- such as car-, bike- or head-mounted, pro or consumer cameras.
Key to these capabilities is a new, large-scale training dataset of images crowd-sourced from cities around the world via the Mapillary platform.

Our experiments show that OrienterNet substantially outperforms previous works on localization in driving scenarios and vastly improves its accuracy in AR use cases when applied to data recorded by Aria glasses.
We believe that our approach constitutes a significant step towards continuous, large scale, on-device localization for AR and robotics.

\begin{table}[t]
\centering
\footnotesize{\begin{tabular}{lccc}
\toprule
Map type & \makecell{SfM\\SLAM} & \makecell{Satellite\\images} & \makecell{OpenStreetMap\\\b{(our work)}}\\
\midrule
What? & \makecell{3D points\\+features} & \makecell{pixel\\intensity} & \makecell{polygons, \\lines, points}\\
Explicit geometry? & 3D & \xmark & 2D\\
Visual appearance? & \cmark & \cmark & \xmark\\
Freely available & \xmark & \xmark & \cmark\\
Storage for \SI{1}{\square\kilo\meter} & \SI{42}{\giga\byte} & \SI{75}{\mega\byte} & \SI{4.8}{\mega\byte}\\
Size reduction vs SfM & - & 550$\times$ & 8800$\times$\\
\bottomrule
\end{tabular}
}
\caption{\textbf{Types of maps for visual localization.}
Planimetric maps from OpenStreetMap consist of polygons and lines with metadata.
They are publicly available for free and do not store sensitive appearance information, as opposed to satellite images and 3D maps built with SfM.
They are also compact: a large area can be downloaded and stored on a mobile device.
We show that they encode sufficient geometric information for accurate 3-DoF localization.
}%
\label{tab:map-types}
\end{table}

\begin{figure}[t]
    \centering
    \includegraphics[width=\linewidth]{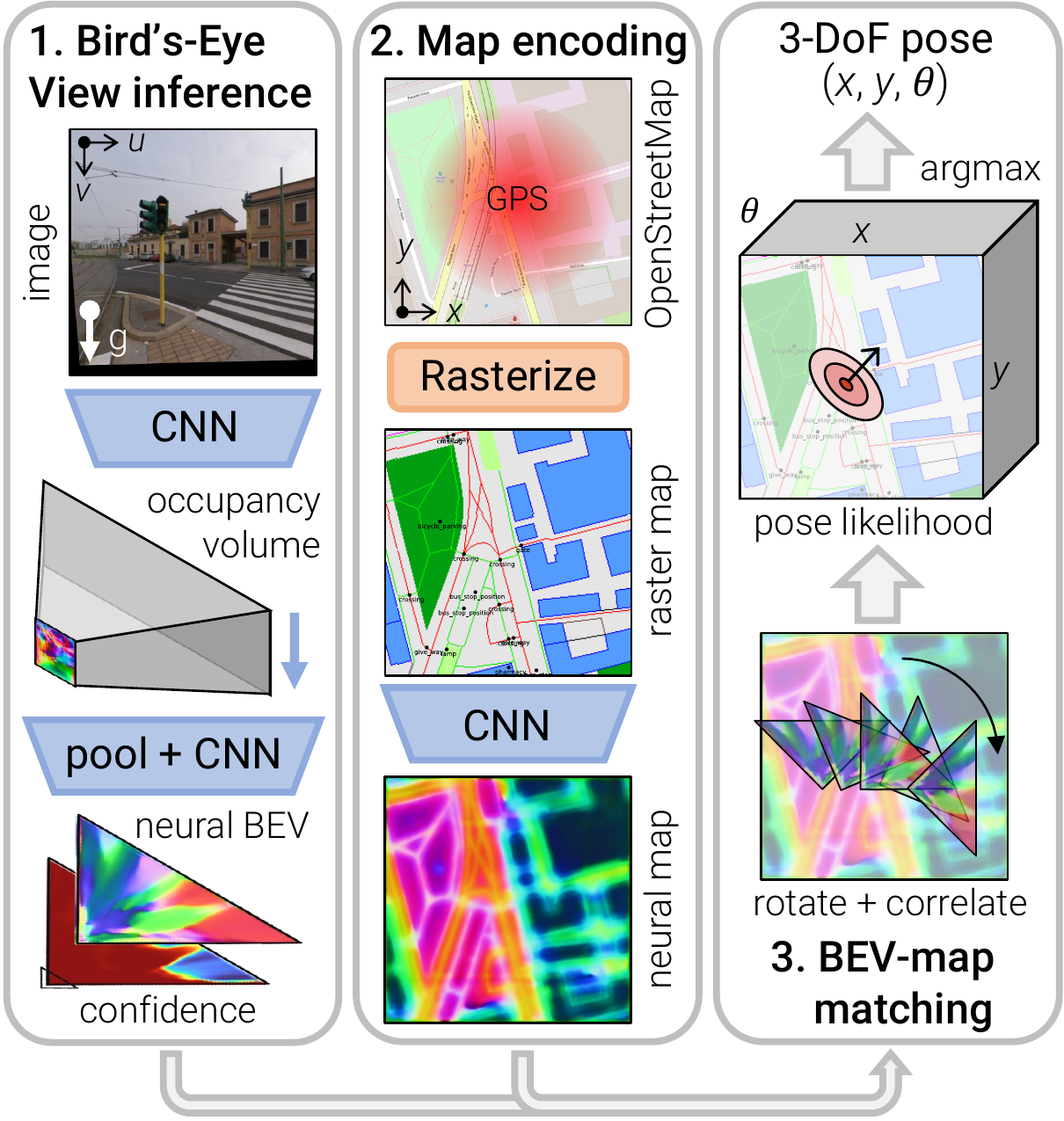}
    \caption{\textbf{OrienterNet architecture.}
    1) From an input image $\*I$ that is gravity-aligned, we infer a mental map of the scene as a neural Bird's-Eye View (BEV) $\*T$ with confidence $\*C$.
    2) From a coarse GPS prior location $\*\xi_\text{prior}$, we query OpenStreetMap and compute a neural map~$\*F$.
    3) Matching the BEV against the map yields a probability volume $\*P$ over 3-DoF camera poses.
    OrienterNet is trained end-to-end from pose supervision only.
    }%
    \label{fig:architecture}%
\end{figure}

\section{Related work}
We can localize an image in the world using several types of map representations: 3D maps built from ground images, 2D overhead satellite images, or simpler planimetric maps from OpenStreetMap.
\Cref{tab:map-types} summarizes their differences.

\PAR{Mapping with ground-level images} is the most common approach to date.
Place recognition via image retrieval provides a coarse localization given a set of reference images~\cite{binarybow,torii201524,vlad,arandjelovic2016netvlad}. 
To estimate centimeter-accurate 6-DoF poses, algorithms based on feature matching require 3D maps~\cite{irschara2009structure,sattler2012improving,Lynen2020IJRR,sarlin2019coarse}. 
These are composed of sparse point clouds, which are commonly built with Structure-from-Motion (SfM)~\cite{snavely2006photo,frahm2010building,agarwal2011building,schoenberger2016sfm,moulon2016openmvg,lindenberger2021pixsfm} from sparse points matched across multiple views~\cite{lowe2004distinctive,bay2006surf,rublee2011orb}. 
The pose of a new query image is estimated by a geometric solver~\cite{Kneip2011CVPR,Bujnak08CVPR,Haralick94IJCV} from correspondences with the map. 
While some works~\cite{Zeisl_2015_ICCV,svarm2017city} leverage additional sensor inputs, such as a coarse GPS location, gravity direction, and camera height, recent localization systems are highly accurate and robust mostly thanks to learned features~\cite{superpoint,sarlin2020superglue,revaud2019r2d2,dusmanu2019d2,tyszkiewicz2020disk}. 

This however involves 3D maps with a large memory footprint as they store dense 3D point clouds with high-dimensional visual descriptors. 
There is also a high risk of leaking personal data into the map. 
To mitigate this, some works attempt to compress the maps~\cite{cao2014minimal,Lynen2020IJRR,Camposeco_2019_CVPR} or use privacy-preserving representations for the scene appearance~\cite{Dusmanu2021Privacy,Ng_2022_CVPR,zhou2022geometry} or geometry~\cite{SpecialeCVPR2019,specialeiccv2019}. 
These however either degrade the accuracy significantly or are easily reverted~\cite{pittaluga2019revealing}.

\PAR{Localization with overhead imagery} reduces the problem to estimating a 3-DoF pose by assuming that the world is mostly planar and that the gravity direction is often given by ubiquitous onboard inertial sensors. 
A large body of work focuses on cross-view ground-to-satellite localization.
While more compact than 3D maps, satellite images are expensive to capture, generally not free, and still heavy to store at high resolution.
Most approaches only estimate a coarse position through patch retrieval~\cite{hu2018cvmnet,NEURIPS2019_ba2f0015,shi2020optimal,zhu2021vigor}. 
In addition, works that estimate an orientation are not accurate~\cite{shi2020looking,shi2022beyond,xia2022visual}, yielding errors of over several meters. 

Other works rely on sensors that directly provide 3D metric information, such as 2D intensity maps from Lidar~\cite{pmlr-v87-barsan18a,ma2019exploiting} or radar~\cite{pmlr-v100-barnes20a,tang2021ijrr}. 
They all perform template matching between 2D map and sensor overhead views, which is both accurate and robust,
but require expensive specialized sensors, unsuitable for consumer AR applications. 
Our work shows how monocular visual priors can substitute such sensors to perform template matching from images only.

\PAR{Planimetric maps} discard any appearance and height information to retain only the 2D location, shape and type of map elements.
OSM is a popular platform for such maps as it is free and available globally.
Given a query area, its open API exposes a list of geographic features as polygons with metadata, including fine-grained semantic information with over a thousand different object types.
Past works however design detectors for a single or few semantic classes, which lacks robustness. 
These include building outlines~\cite{armagan2017accurate,armagan2017learning,cham2010estimating,chu2014gps,david2011Orientation,vojir2020Efficient,vysotska2017improving}, road contours~\cite{ruchti2015Localization,Floros2013OpenStreetSLAM} or intersections~\cite{ma2017find,  panphattarasap2018automated,yan2019global}, lane markings~\cite{guo2021coarse,pauls2020monocular}, street furniture~\cite{weng2021semantic,castaldo2015semantic}, or even text~\cite{hong2019textplace}.

Recent works leverage more cues by computing richer representations from map tiles using end-to-end deep networks~\cite{samano2020you,zhou2021efficient}. 
They estimate only a coarse position as they retrieve map tiles with global image descriptors. 
In indoor scenes, floor plans are common planimetric maps used by existing works~\cite{howard2021lalaloc,min2022laser}.
They require height or visibility information that is typically not available for outdoor spaces.
Our approach yields a significant step up in accuracy and robustness over all previous works by combining the constraints of projective geometry with the expressivity of end-to-end learning, leveraging all semantic classes available in OSM.

\section{Localizing single images in 2D maps}

\PAR{Problem formulation:} 
In a typical localization scenario, we aim to estimate the absolute 6-DoF pose of an image in the world. 
Under realistic assumptions, we reduce this problem to estimating a 3-DoF pose $\*\xi = (x,y,\theta)$ consisting of a location $(x,y) \in \mathbb{R}^2$ and heading angle $\theta\in(-\pi,\pi]$.
Here we consider a topocentric coordinate system whose $x$-$y$-$z$ axes correspond to the East-North-vertical directions. 

First, we can easily assume to know the direction of the gravity, an information that humans naturally possess from their inner ear and that can be estimated by the inertial unit embedded in most devices. 
We also observe that our world is mostly planar and that the motion of people and objects in outdoor spaces is mostly restricted to 2D surface. 
The precise height of the camera can always be estimated as the distance to the ground in a local SLAM reconstruction. 

\PAR{Inputs:}
We consider an image $\*I$ with known pinhole camera calibration. 
The image is rectified via a homography computed from the known gravity such that its roll and tilt are zero -- its principal axis is then horizontal. 
We are also given a coarse location prior $\*\xi_\text{prior}$. 
This can be a noisy GPS position or a previous localization estimate and can be off by over 20 meters. 
This is a realistic assumption for a consumer-grade sensor in a multi-path environment like a urban canyon.

The map data is queried from OSM as a square area centered around $\*\xi_\text{prior}$ and whose size depends on how noisy the prior is. 
The data consists of a collection of polygons, lines, and points, each of a given semantic class and whose coordinates are given in the same local reference frame. 

\PAR{Overview -- \Cref{fig:architecture}:} 
OrienterNet consists of three modules: 
1)~The image-CNN extracts semantic features from the image and lifts them to an orthographic Bird's-Eye View (BEV) representation $\*T$ by inferring the 3D structure of the scene.
2)~The OSM map is encoded by the map-CNN into a neural map $\*F$ that embeds semantic and geometric information. 
3)~We estimate a probability distribution over camera poses $\*\xi$ by exhaustively matching the BEV against the map.

\begin{figure}[t]
    \centering
    \includegraphics[width=0.9\linewidth]{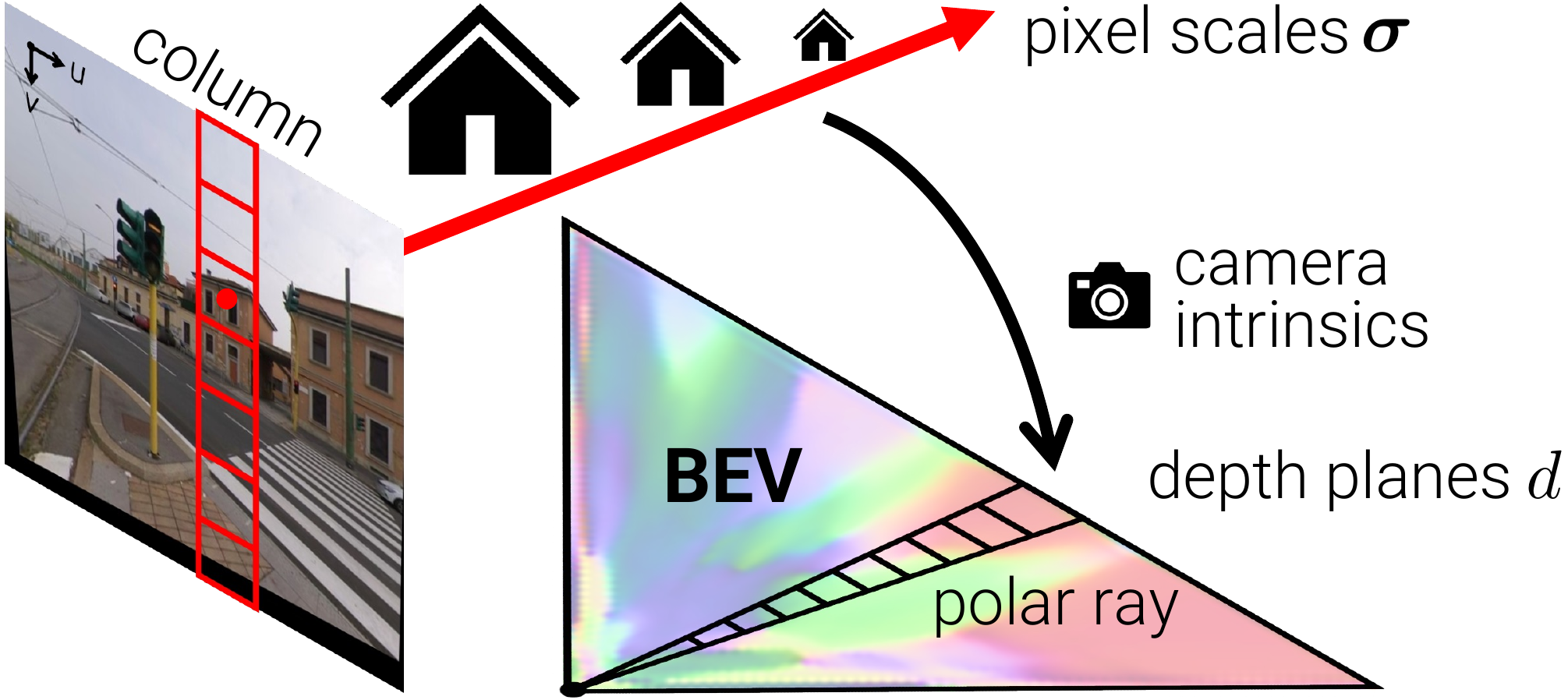}
    \caption{\textbf{OrienterNet predicts a pixel-wise distribution over scales}
    that are mapped to depths with the known camera calibration.
    }%
    \label{fig:bev}%
\end{figure}

\subsection{Neural Bird's-Eye View inference}

\PAR{Overview:}
From a single image $\*I$, we infer a BEV representation $\*T \in \mathbb{R}^{L\times D \times N}$ distributed on a $L{\times}D$ grid aligned with the camera frustum and composed of $N$-dimensional features.
Each feature on the grid is assigned a confidence, yielding a matrix $\*C\in[0,1]^{L\times D}$.
This BEV is akin to a mental map that humans infer from their environment when self-localizing in an overhead map~\cite{O'Keefe1978,lobben2007navigational}.

Cross-modal matching between the image and the map requires extracting semantic information from visual cues.
It has been shown that monocular depth estimation can rely on semantic cues~\cite{antequera2020mapillary} and that both tasks have a beneficial synergy~\cite{hoyer2021three,lao2022does}.
We thus rely on monocular inference to lift semantic features to the BEV space.
Following past works that tackle semantic tasks~\cite{roddick2020predicting,philion2020lift,saha2022translating}, we obtain the neural BEV in two steps:
i) we transfer image features to a polar representation by mapping image columns to polar rays, and ii) we resample the polar grid into a Cartesian grid (\cref{fig:bev}).

\PAR{Polar representation:}
A CNN $\Phi_\text{image}$ first extracts a $U{\times}V$ feature map $\*X\in\mathbb R^{U\times V\times N}$ from the image.
We consider $D$ depth planes sampled in front of the camera with a regular interval $\Delta$, \ie with values $\left\{i\cdot\Delta | i \in \{1 \dots D\}\right\}$.
Since the image is gravity-aligned, each of the $U$ columns in $\*X$ corresponds to a vertical plane in the 3D space.
We thus map each column to a ray in the $U{\times}D$ polar representation $\bar{\*X}\in\mathbb{R}^{U\times D\times N}$.
We do so by predicting, for each polar cell $(u,d)$, a probability distribution $\*\alpha_{u,d} \in [0,1]^V$ over the pixels in the corresponding image column:
\begin{equation}
\bar{\*X}_{u,d} = \sum_v \*\alpha_{u,d,v}\*X_{u,v}\enspace.
\end{equation}

Instead of directly regressing the distribution $\*\alpha$ over depths, we regress a distribution $\*S$ over \emph{scales} that are independent from the camera calibration parameters.
The scale is the ratio of object sizes in the 3D world and in the image~\cite{antequera2020mapillary}
and is equal to the ratio of the focal length $f$ and depth.
We consider a set of $S$ log-distributed scales
\begin{equation}
\*\sigma=\left\{\sigma_\text{min} \left(\sigma_\text{max}/\sigma_\text{min}\right)^{i/S} | i\in\{0\dots S\}\right\}
\enspace.
\end{equation}
$\Phi_\text{image}$ also predicts, for each pixel $(u,v)$, a score vector $\*S_{u,v}\in\mathbb{R}^S$ whose elements correspond to the scale bins $\*\sigma$.
We then obtain the distribution $\*\alpha_{u,d}$ for each depth bin $d$ as
\begin{equation}
\*\alpha_{u,d,v} = \underset{v}{\operatorname*{softmax}}\left(\*S_{u,v}\left[\nicefrac{f}{d\cdot \Delta}\right]\right)\enspace,
\end{equation}
where $[\cdot]$ denotes the linear interpolation.

This formulation is equivalent to an attention mechanism from polar rays to image columns with scores resampled from linear depths to log scales.
When the scale is ambiguous and difficult to infer, visual features are spread over multiple depths along the ray but still provide geometric constraints for well-localized map points~\cite{Larsson_2021_ICCV}.
Works tailored to driving scenarios~\cite{roddick2020predicting,philion2020lift,saha2022translating} consider datasets captured by cameras with identical models and directly regress $\*\alpha$.
They therefore encode the focal length in the network weights, learning the mapping from object scale to depth.
Differently, our formulation can generalize to arbitrary cameras at test time by assuming that the focal length is an input to the system.

\PAR{BEV grid:}
We map the polar features to a Cartesian grid of size $L{\times}D$ via linear interpolation along the lateral direction from $U$ polar rays to $L$ columns spaced by the same interval~$\Delta$.
The resulting feature grid is then processed by a small CNN $\Phi_\text{BEV}$ that outputs the neural BEV $\*T$ and confidence~$\*C$.

\subsection{Neural map encoding}
We encode the planimetric map into a $W{\times}H$ neural map $\*F \in \mathbb{R}^{W{\times}H{\times}N}$
that combines geometry and semantics.

\PAR{Map data:}
OpenStreetMap elements are defined, depending on their semantic class, as polygonal areas, multi-segment lines, or single points.
Examples of areas include building footprints, grass patches, parking lots; lines include road or sidewalk center lines, building outlines; points include trees, bus stops, shops, etc.
\supp~\ref{sec:supp:osm} lists all classes.
The accurate positioning of these elements provides geometric constraints necessary for localization, while their rich semantic diversity helps disambiguate different poses.

\PAR{Preprocessing:}
We first rasterize the areas, lines, and points as a 3-channels image with a fixed ground sampling distance~$\Delta$, e.g. \SI[per-mode=symbol,mode=text]{50}{\centi\meter\per\pixel}.
This representation is more informative and accurate than the naive rasterization of human-readable OSM tiles performed in previous works~\cite{samano2020you,zhou2021efficient}.

\PAR{Encoding:} We associate each class with an $N$-dimensional embedding that is learned, yielding a $W{\times}H{\times}3N$ feature map.
It is then encoded into the neural map $\*F$ by a CNN $\Phi_\text{map}$, which extracts geometric features useful for localization.
$\*F$~is not normalized as we let $\Phi_\text{map}$ modulate its norm as importance weight in the matching.
Examples in \cref{fig:qualitative-main} reveal that $\*F$ often looks like a distance field where we can clearly recognize distinctive features like corners or adjoining boundaries of buildings.

$\Phi_\text{map}$ also predicts a unary location prior $\*\Omega\in\mathbb{R}^{W\times H}$ for each cell of the map.
This score reflects how likely an image is to be taken at each location.
We rarely expect images to be taken in, for example, rivers or buildings.

\begin{figure*}[t]
    \centering
    \input{figures/qualitative_v3}
    \vspace{0.2mm}%
    \caption{\textbf{We train a single model that generalizes well across many datasets.}
    OrienterNet handles different cameras, street-level viewpoints, and unseen cities and is thus suitable for both AR and robotics.
    Overlayed on the input maps, the single-image predictions (black~arrow \raisebox{-2pt}{\includegraphics[height=9pt]{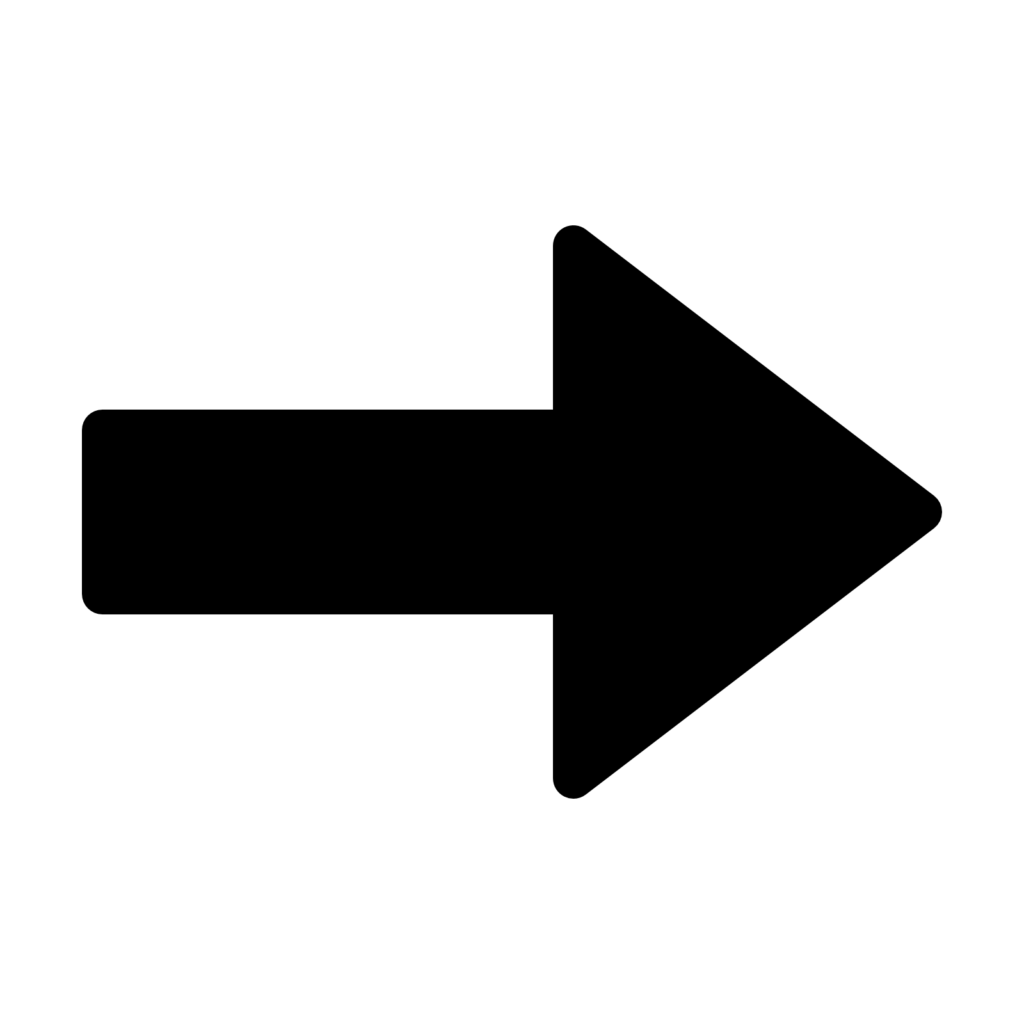}}) are close to the ground truth (red arrow) and more accurate than the noisy GPS (blue dot \cbullet{blue}).
    The model effectively leverages building corners and boundaries, crosswalks, sidewalks, road intersections, trees, and other common urban objects.
    Likelihood maps can be multi-modal in scenes with repeated elements (last two rows) - we show the predicted orientations at local maxima as small arrows.
    }%
    \label{fig:qualitative-main}%
\end{figure*}

\subsection{Pose estimation by template matching}

\PAR{Probability volume:}
We estimate a discrete probability distribution over camera poses $\*\xi$.
This is interpretable and fully captures the uncertainty of the estimation. As such, the distribution is multimodal in ambiguous scenarios. 
\Cref{fig:qualitative-main} shows various examples.
This makes it easy to fuse the pose estimate with additional sensors like GPS.
Computing this volume is tractable because the pose space has been reduced to 3 dimensions. 
It is discretized into each map location and $K$ rotations sampled at regular intervals.

This yields a $W{\times}H{\times}K$ probability volume $\*P$ such that $P(\*\xi|\*I,\text{map},\*\xi_\text{prior})=\*P[\*\xi]$.
It is the combination of an image-map matching term $\*M$ and the location prior $\*\Omega$:
\begin{equation}
\*P =\operatorname*{softmax} \left(\*M + \*\Omega\right)
\enspace.
\end{equation}
$\*M$ and $\*\Omega$ represent image-conditioned and image-independent un-normalized log scores.
$\*\Omega$ is broadcasted along the rotation dimension and softmax normalizes the probability distribution.

\PAR{Image-map matching:}
Exhaustively matching the neural map $\*F$ and the BEV $\*T$ yields a score volume $\*M$.
Each element is computed by correlating $\*F$ with $\*T$ transformed by the corresponding pose as
\begin{equation}
\*M[\*\xi] = \frac{1}{UZ}\sum_{\*p\in(U\times Z)}\*F[\*\xi(\*p)]^\top\left(\*T\odot\*C\right)[\*p] 
\enspace,
\end{equation}
where
$\*\xi(\*p)$ transforms a 2D point $\*p$ from BEV to map coordinate frame.
The confidence $\*C$ masks the correlation to ignore some parts of the BEV space, such as occluded areas.
This formulation benefits from an efficient implementation by rotating $\*T$ $K$ times and performing a single convolution as a batched multiplication in the Fourier domain~\cite{pmlr-v87-barsan18a,pmlr-v100-barnes20a}.

\PAR{Pose inference:}
We estimate a single pose by maximum likelihood:
$\*\xi^* = \operatorname*{argmax}_{\*\xi} P(\*\xi|\*I,\text{map},\*\xi_\text{prior})$.
When the distribution is mostly unimodal, we can obtain a measure of uncertainty as the covariance of $\*P$ around $\*\xi^*$~\cite{pmlr-v100-barnes20a}.

\section{Sequence and multi-camera localization}
Single-image localization is ambiguous in locations that exhibit few distinctive semantic elements or repeated patterns.
Such challenge can be disambiguated by accumulating additional cues over multiple views when their relative poses are known.
These views can be either sequences of images with poses from VI SLAM or simultaneous views from a calibrated multi-camera rig.
\Cref{fig:qualitative-sequence} shows an example of such difficult scenario disambiguated by accumulating predictions over time.
Different frames constrain the pose in different directions, \eg before and after an intersection.
Fusing longer sequences yields a higher accuracy 
(\cref{fig:sequence-aria}).

Let us denote $\*\xi_i$ the unknown absolute pose of view $i$ and $\*{\hat{\xi}}_{ij}$ the known relative pose from view $j$ to $i$.
For an arbitrary reference view $i$, we express the joint likelihood over all single-view predictions as
\begin{equation}
P(\*\xi_i|\{\*I_j\},\text{map}) = \prod_k P(\*\xi_i\oplus\*{\hat{\xi}}_{ij}|\*I_j,\text{map})\enspace,
\end{equation}
where $\oplus$ denotes the pose composition operator.
This is efficiently computed by warping each probability volume $\*P_j$ to the reference frame $i$.
We can also localize each image of a continuous stream
via iterative warping and normalization, like in the classical Markov localization~\cite{SimmonsK95,burgard1996estimating}.

\begin{figure}[t]
    \centering
    \input{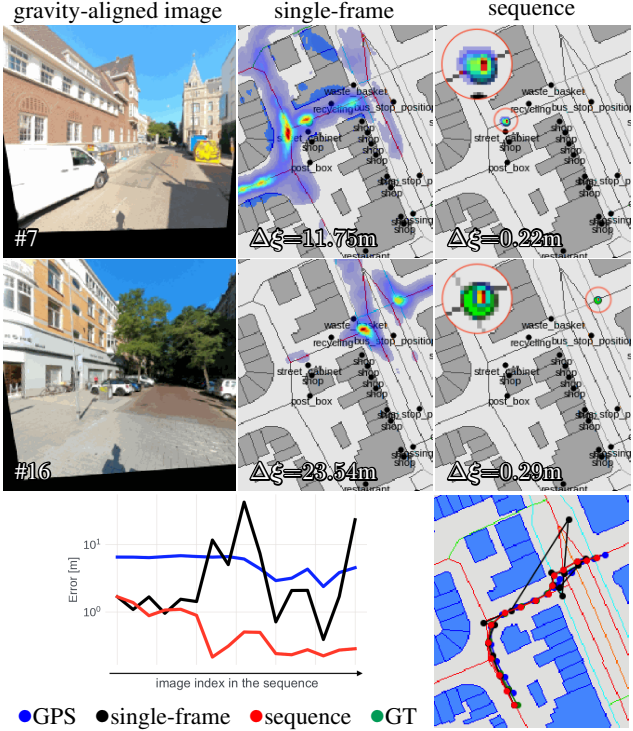}
    \caption{\textbf{Multi-frame fusion resolves ambiguities.}
    Semantic elements visible in a single image are often not sufficient to fully disambiguate the camera pose.
    Fusing the predictions over multiple frames collapses the multi-modal likelihood map to a single mode with high accuracy, yielding here a final error of less than 30cm.
    }%
    \label{fig:qualitative-sequence}%
\end{figure}

\section{Training a single strong model}

\PAR{Supervision:}
OrienterNet is trained in a supervised manner from pairs of single images and ground truth (GT) poses.
The architecture is differentiable and all components are trained simultaneously by back-propagation.
We simply maximize the log-likelihood of the ground truth pose~$\*\xi$: $\text{Loss} = -\log P(\*\xi|\*I,\text{map},\*\xi_\text{prior}) = -\log \*P[\*\xi]$.
The tri-linear interpolation of $\*P$ provides sub-pixel supervision.

\begin{figure}[t]
    \centering
    \includegraphics[width=\linewidth]{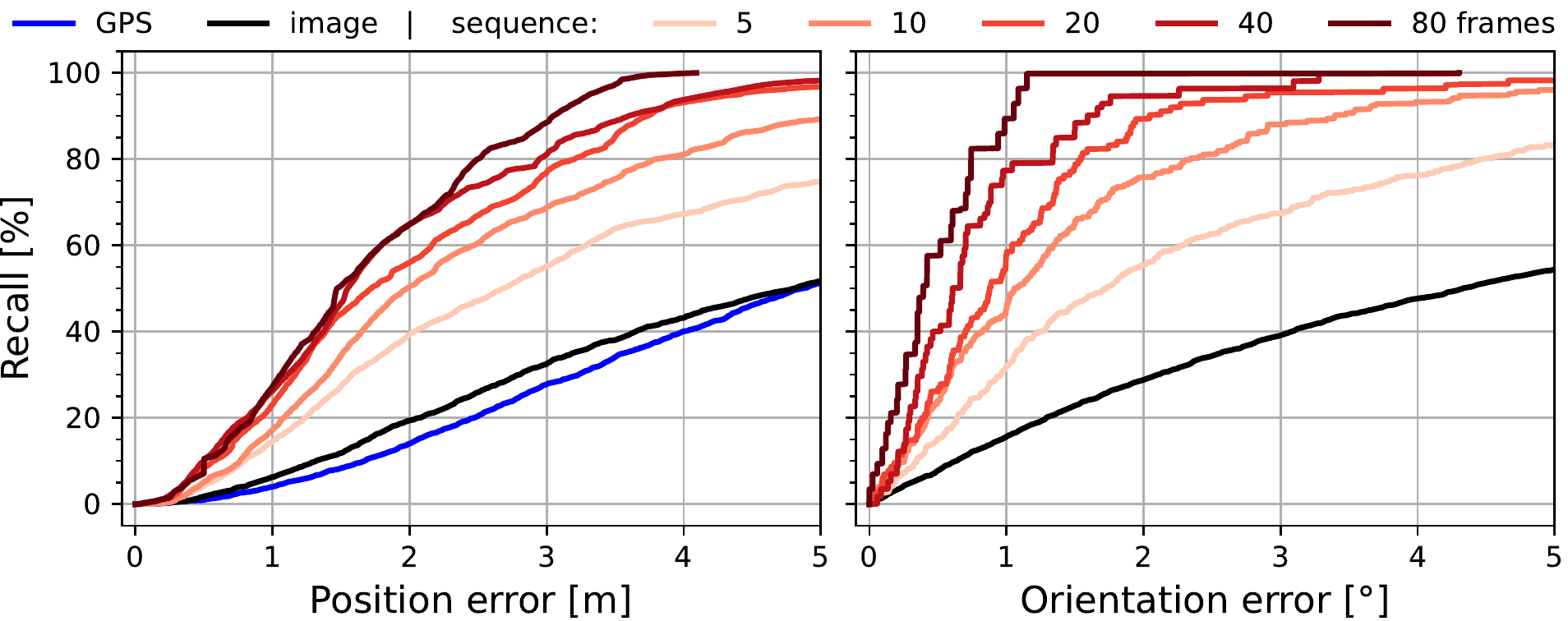}%
    \caption{\textbf{With AR data, sequence localization boosts the recall}, 
    which increases as we fuse information from additional frames.
    }%
    \label{fig:sequence-aria}%
\end{figure}

\begin{table}[t]
\centering
\footnotesize{\setlength{\tabcolsep}{3.5pt}
\begin{tabular}{lcccccc}
\toprule
\multirow{2}{*}{Model architecture}
& \multicolumn{3}{c}{Position R@Xm}
& \multicolumn{3}{c}{Orientation R@X\degree}\\
\cmidrule(lr){2-4}
\cmidrule(lr){5-7}
& 1m & 3m & 5m & 1\degree & 3\degree & 5\degree\\
\midrule
Retrieval (a) & \02.02 & 15.21 & 24.21 & \04.50 & 18.61 & 32.48\\
Refinement (b) & \08.09 & 26.02 & 35.31 & 14.92 & 36.87 & 45.19\\
OrienterNet - planar (c) & 14.28 & 44.59 & 56.08 & 20.43 & 50.34 & 64.30\\
\b{OrienterNet - full} & \b{15.78}	& \b{47.75} & \b{58.98} & \b{22.14} & \b{52.56} & \b{66.32}\\
\bottomrule
\end{tabular}}
\caption{\textbf{OrienterNet outperforms existing architectures}, which include:
a)~map tile retrieval by matching global embeddings~\cite{xia2022visual,samano2020you},
b)~featuremetric refinement~\cite{shi2022beyond} from an initial pose, and
c)~OrienterNet assuming a planar scene~\cite{shi2022beyond} instead of inferring monocular depth.
We report the position and orientation recall (R).
}%
\label{tab:architectures}
\end{table}

\PAR{Training dataset:}
We train a single model that generalizes to unseen locations with arbitrary kinds of images.
We collect images from the Mapillary platform, which exposes the camera calibration, noisy GPS measurement, and the 6-DoF pose in a global reference frame, obtained with a fusion of SfM and GPS.
The resulting \emph{Mapillary Geo-Localization} (MGL) dataset includes 760k images from 12 cities in Europe and the US, captured by cameras that are handheld or mounted on cars or bikes, with GT poses and OSM data.
Models trained on MGL generalize well to other datasets thanks to the diversity of cameras, locations, motions, and maps.
All images are publicly available under a CC-BY-SA license via the Mapillary API.
We believe that this dataset will significantly facilitate research on visual geo-localization.

\begin{table*}[t]
\centering
\footnotesize{\setlength{\tabcolsep}{7.5pt}
\begin{tabular}{llcccccccccc}
\toprule
\multirow{2}{*}{Map} & \multirow{2}{*}{Approach} & \multirow{2}{*}[-.3em]{\makecell{Training\\dataset}}
& \multicolumn{3}{c}{Lateral R@Xm}
& \multicolumn{3}{c}{Longitudinal R@Xm}
& \multicolumn{3}{c}{Orientation R@X\degree}\\
\cmidrule(lr){4-6}
\cmidrule(lr){7-9}
\cmidrule(lr){10-12}
&&& 1m & 3m & 5m & 1m & 3m & 5m & 1\degree & 3\degree & 5\degree\\
\midrule
Satellite & DSM~\cite{shi2020looking} & KITTI & 10.77 & 31.37 & 48.24 & \03.87 & 11.73 & 19.50 & \03.53 & 14.09 & 23.95\\
& VIGOR~\cite{zhu2021vigor} & KITTI & 17.38 & 48.20 & 70.79 & \04.07 & 12.52 & 20.14 & - & - & -\\
& refinement~\cite{shi2022beyond} & KITTI & 27.82 & 59.79 & 72.89 & \05.75 & 16.36 & 26.48 & 18.42 & 49.72 & 71.00\\
\midrule
OpenStreetMap & retrieval~\cite{xia2022visual,samano2020you} & MGL & 37.47 & 66.24 & 72.89 & \05.94 & 16.88 & 26.97 & \02.97 & 12.32 & 23.27\\
& refinement~\cite{shi2022beyond} & MGL & 
50.83 & 78.10 & 82.22 & 17.75 & 40.32 & 52.40 & 31.03 & 66.76 & 76.07\\
& \textbf{OrienterNet} (a) & MGL & 53.51 & 88.85 & 94.47 & 26.25 & 59.84 & 70.76 & 34.26 & 73.51 & 89.45\\
& \textbf{\raisebox{.4ex}{$\drsh$} + sequence} (b) & MGL & \b{79.71} & \b{97.44} & \b{98.67} & \b{55.21} & \b{95.27} & \b{99.51} & \b{77.87} & \b{97.76} & \b{100.}\\
\cmidrule{2-12}
& \textbf{OrienterNet} (c) & KITTI & 51.26 & 84.77 & 91.81 & 22.39 & 46.79 & 57.81 & 20.41 & 52.24 & 73.53\\
& \textbf{OrienterNet} (d) & MGL+KITTI & 65.91 & 92.76 & 96.54 & 33.07 & 65.18 & 75.15 & 35.72 & 77.49 & 91.51\\
\bottomrule
\end{tabular}}
\caption{\textbf{Localization in driving scenarios with the KITTI dataset.}
a)~When trained on our MGL dataset, OrienterNet yields a higher localization recall than existing approaches based on both satellite imagery and OpenStreetMap, in terms of both orientation and lateral and longitudinal positional errors, 
b)~Fusing predictions from sequences of 20 seconds boosts the recall.
c)~Training on KITTI outperforms other approaches trained on KITTI but is inferior to training on MGL. This demonstrates the excellent zero-shot capability of OrienterNet and the value of MGL.
d) Pre-training on MGL and fine-tuning on KITTI achieves the best single-image performance.
}%
\label{tab:eval-kitti}
\end{table*}

\PAR{Implementation:}
$\Phi_{\text{image}}$ and $\Phi_{\text{map}}$ are U-Nets with ResNet-101 and VGG-16 encoders.
$\Phi_{\text{BEV}}$ has 4 residual blocks.
We use $S{=}32$ scale bins, $K{=}512$ rotations.
The BEV has size $L{\times}D{=}\SI{32}{}{\times}\SI{32}{\meter}$ with resolution $\Delta{=}\SI{50}{\centi\meter}$.
For training, we render maps $W{\times}H{=}\SI{128}{}{\times}\SI{128}{\meter}$ centered around points randomly sampled within \SI{32}{\meter} of the GT pose.
Localizing in such map takes \SI{94}{\milli\second} on an NVIDIA RTX 2080 GPU, with \SI{37}{\milli\second} for the BEV inference
and \SI{51}{\milli\second} for the matching.

\section{Experiments}
We evaluate our single model for localization in the context of both driving and AR.
\Cref{fig:qualitative-main} shows qualitative examples, while \cref{fig:qualitative-sequence} illustrates the effectiveness of multi-frame fusion.
Our experiments show that:
1) OrienterNet is more effective than existing deep networks for localization with 2D maps;
2) Planimetric maps help localize more accurately than overhead satellite imagery;
3) OrienterNet is significantly more accurate than an embedded consumer-grade GPS sensor when considering multiple views.

\subsection{Validating design decisions}
\label{sec:validation}

\PAR{Setup:}
We evaluate the design of OrienterNet on the validation split of our MGL dataset.
This ensures an identical distribution of cameras, motions, viewing conditions, and visual features as the training set.
We report recall of positions and rotation errors at the three thresholds 1/3/5m and 1/3/5$\degree$.

\PAR{Comparing model architectures:}
We compare OrienterNet to alternative architectures trained on the same dataset:
a)~\underline{Map retrieval}~\cite{xia2022visual} replaces the BEV inference and matching by a correlation of the neural map and with a global image embedding.
We predict a rotation by considering 4 different neural maps for the N-S-E-W directions.
This formulation also regresses a probability volume and is trained identically to OrienterNet.
It mimics the retrieval of densely-sampled map patches~\cite{samano2020you} but is significantly more efficient and practical.
b)~\underline{Featuremetric refinement}~\cite{shi2022beyond,sarlin21pixloc} updates an initial pose by warping a satellite view to the image assuming that the scene is planar, at a fixed height, and gravity-aligned.
We replace the satellite view by an OSM map tile.
This formulation requires an initial orientation (during both training and testing), which we sample within 45$\degree$ of the ground truth.
c)~\underline{OrienterNet (planar)} replaces the occupancy by warping the image features with a homography as in~\cite{shi2022beyond}.

\PAR{Analysis -- \Cref{tab:architectures}:}
OrienterNet is significantly more accurate than all baselines at all position and rotation thresholds.
a)~Map retrieval disregards any knowledge of projective geometry and performs mere recognition without any geometric constraint.
b)~Featuremetric refinement converges to incorrect locations when the initial pose is inaccurate.
c)~Inferring the 3D geometry of the scene is more effective than assuming that it is planar.
This justifies our design decisions.

\PAR{Model interpretability:}
We visualize in \cref{fig:introspection} multiple internal quantities that help us understand the predictions.

\subsection{Application: autonomous driving}
\label{sec:driving}

\PAR{Dataset:}
We consider the localization in driving scenarios with the KITTI dataset~\cite{geiger2013vision},
following the closest existing setup~\cite{shi2022beyond}. 
To evaluate the zero-shot performance, we use their \emph{Test2} split, which does not overlap with the KITTI and MGL training sets.
Images are captured by cameras mounted on a car driving in urban and residential areas and have GT poses from RTK. 
We augment the dataset with OSM maps.

\PAR{Setup:}
We compute the position error along directions perpendicular (lateral) and parallel (longitudinal) to the viewing axis~\cite{shi2022beyond} since the pose is generally less constrained along the road.
We report the recall at 1/3/5m and 1/3/5$\degree$.
The original setup~\cite{shi2022beyond} assumes an accurate initial pose randomly sampled within $\pm$20m and $\pm$10$\degree$ of the GT.
OrienterNet does not require such initialization but only a coarse position-only prior.
For fair comparisons, we nevertheless restrict the pose space to the same interval centered around the initial pose.
We render $\SI{64}{\meter}{\times}\SI{64}{\meter}$ map tiles and resize the images such that their focal length matches the median of MGL.

\PAR{Baselines:}
We report approaches based on satellite maps and trained by~\cite{shi2022beyond} on KITTI.
VIGOR~\cite{zhu2021vigor} and DSM~\cite{shi2020looking} both perform patch retrieval with global descriptors but respectively estimate an additional position offset or the orientation.
We also evaluate the featuremetric refinement~\cite{shi2022beyond,sarlin21pixloc} and baselines based on OSM maps, described in \cref{sec:validation}.
As each scene is visited by a single trajectory, we cannot evaluate approaches based on 3D maps and image matching.

\PAR{Results:}
\Cref{tab:eval-kitti} (a-b) shows that OrienterNet outperforms all existing approaches based on both satellite and OSM maps, in all metrics.
OrienterNet exhibits remarkable zero-shot capabilities as it outperforms approaches trained on KITTI itself.
The evaluation also demonstrates that planimetric maps yield better localization, as retrieval and refinement approaches based on them outperform those based on satellite images.
The recall at 3m/3\degree is saturated to over 95\% by fusing the predictions from sequences of only 20 seconds.

\PAR{Generalization:}
\Cref{tab:eval-kitti} (c-d) shows that training OrienterNet solely on KITTI results in overfitting, as the dataset is too small to learn rich semantic representations.
Our larger MGL dataset alleviates this issue and enables cross-modal learning with rich semantic classes.
Pre-training on MGL and fine-tuning on KITTI yields the best performance.

\begin{table}[t]
\centering
\footnotesize{\setlength{\tabcolsep}{2.4pt}
\begin{tabular}{cclcccccc}
\toprule
\multirow{2}{*}{City}&\multirow{2}{*}{Setup} & \multirow{2}{*}{Approach} & \multicolumn{3}{c}{Position R@Xm} & \multicolumn{3}{c}{Orientation R@X\degree}\\
\cmidrule(lr){4-6}
\cmidrule(lr){7-9}
&&& 1m & 3m & 5m & 1\degree & 3\degree & 5\degree\\
\midrule

\multirow{5}{0.3cm}{\begin{sideways}Seattle\end{sideways}} & single & GPS & \01.25 & \08.82 & 18.44 & - & - & -\\
&& retrieval~\cite{xia2022visual,samano2020you} & \00.88 & \03.81 & \05.95 & \02.83 & \08.36 & 12.96\\
&& \textbf{OrienterNet} & \03.39 & 14.49 & 23.92 & \06.83 & 20.39 & 30.89\\

\cmidrule{2-9}

 & multi & GPS & \01.76 & \09.2 & 20.48 & \04.18 & 11.01 & 23.36\\
&& \textbf{OrienterNet} & \b{21.88} & \b{61.26} & \b{72.92} & \b{33.86} & \b{72.41} & \b{83.93}\\

\midrule

\multirow{5}{0.3cm}{\begin{sideways}Detroit\end{sideways}} & single & GPS & \03.96 & 27.75 & 51.33 & - & - & -\\
&& retrieval~\cite{xia2022visual,samano2020you} & \03.31 & 19.83 & 36.76 & \06.48 & 18.40 & 28.88\\
&& \textbf{OrienterNet} & \06.26 & 32.41 & 51.76 & 15.53 & 39.06 & 54.41\\

\cmidrule{2-9}

& multi & GPS &  \04.09 & 31.36 & 53.41 & 13.48 & 37.84 & 55.24\\
&& \textbf{OrienterNet} & \b{17.18} & \b{68.77} & \b{89.26} & \b{44.85} & \b{88.04} & \b{96.04}\\
\bottomrule
\end{tabular}}
\caption{\textbf{Localization of head-mounted devices for AR.}
With data from Aria glasses, OrienterNet outperforms the map retrieval baseline and the embedded GPS sensor in both single- and multi-frame settings, in both cities.
Multi-frame fusion does not filter out the high noise of the GPS but strongly benefits our approach.
}%
\label{tab:eval-aria}
\end{table}

\subsection{Application: augmented reality}
We now consider the localization of head-mounted devices for augmented reality (AR).
We show that OrienterNet is more accurate than a typical embedded GPS sensor.

\PAR{Dataset:}
There is no public benchmark that provides geo-aligned GT poses for images captured with AR devices in diverse outdoor spaces.
We thus record our own dataset with Aria~\cite{aria} glasses.
It exhibits patterns typical of AR with noisy consumer-grade sensors and pedestrian viewpoints and motions.
We include two locations: i)~downtown Seattle, with high-rise buildings, and ii)~Detroit, with city parks and lower buildings.
We record several image sequences per city, all roughly following the same loop around multiple blocks.
We record calibrated RGB images and GPS measurements for each and obtain relative poses and gravity direction from an offline proprietary VI SLAM system.
We obtain pseudo-GT global poses by jointly optimizing all sequences based on GPS, VI constraints, and predictions of OrienterNet.

\PAR{Single-frame localization -- \Cref{tab:eval-aria}:}
OrienterNet is consistently more accurate than the GPS, which is extremely noisy in urban canyons like Seattle because of multi-path effects.
The performance is however significantly lower than with driving data (\cref{sec:driving}), which highlights the difficulty of AR-like conditions and the need for further research.

\PAR{Multi-frame:}
We now fuse multiple GPS signals or predictions of OrienterNet over the same temporal interval of 10 consecutive keyframes, using imperfect relative poses from VI SLAM.
The fusion more than doubles the accuracy of OrienterNet but marginally benefits the GPS sensor because of its high, biased noise, especially in Seattle.

\PAR{Limitations:}
Localizing an image or a sequence is challenging when the environments lacks distinctive elements or when they are not registered in the map.
OSM may also be spatially inaccurate.
\Cref{sec:supp:results} shows some failure cases.

\begin{figure}[t]
    \centering
    \input{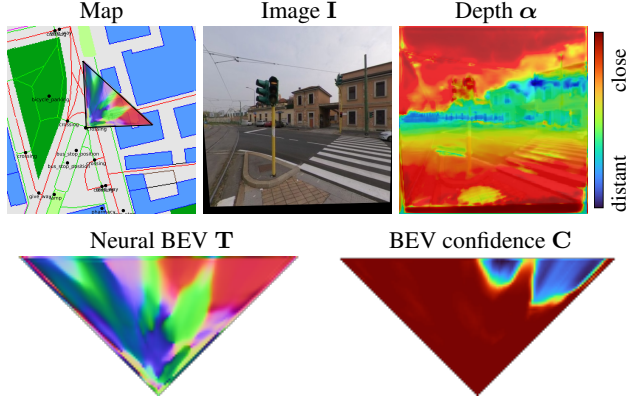}
    \caption{\textbf{End-to-end but interpretable.}
    From only pose supervision, OrienterNet learns to infer the 3D geometry of the scene via the depth planes $\*\alpha$ and the 2D occupancy via the confidence $\*C$.
    }%
    \label{fig:introspection}%
\end{figure}

\section{Conclusion}
OrienterNet is the first deep neural network that can localize an image with sub-meter accuracy within the same 2D planimetric maps that humans use.
OrienterNet mimics the way humans orient themselves in their environment by matching the input map with a mental map derived from visual observations.
Compared to large and expensive 3D maps that machines have so far relied on, such 2D maps are extremely compact and thus finally enable on-device localization within large environments.
OrienterNet is based on globally and freely available maps from OpenStreetMap and can be used by anyone to localize anywhere in the world.

We contribute a large, crowd-sourced training dataset that helps the model generalize well across both driving and AR datasets.
OrienterNet significantly improves over existing approaches for 3-DoF localization, pushing the state of the art by a large margin.
This opens up exciting prospects for deploying power-efficient robots and AR devices that know where they are without costly cloud infrastructures.

\begin{figure*}[ht!]
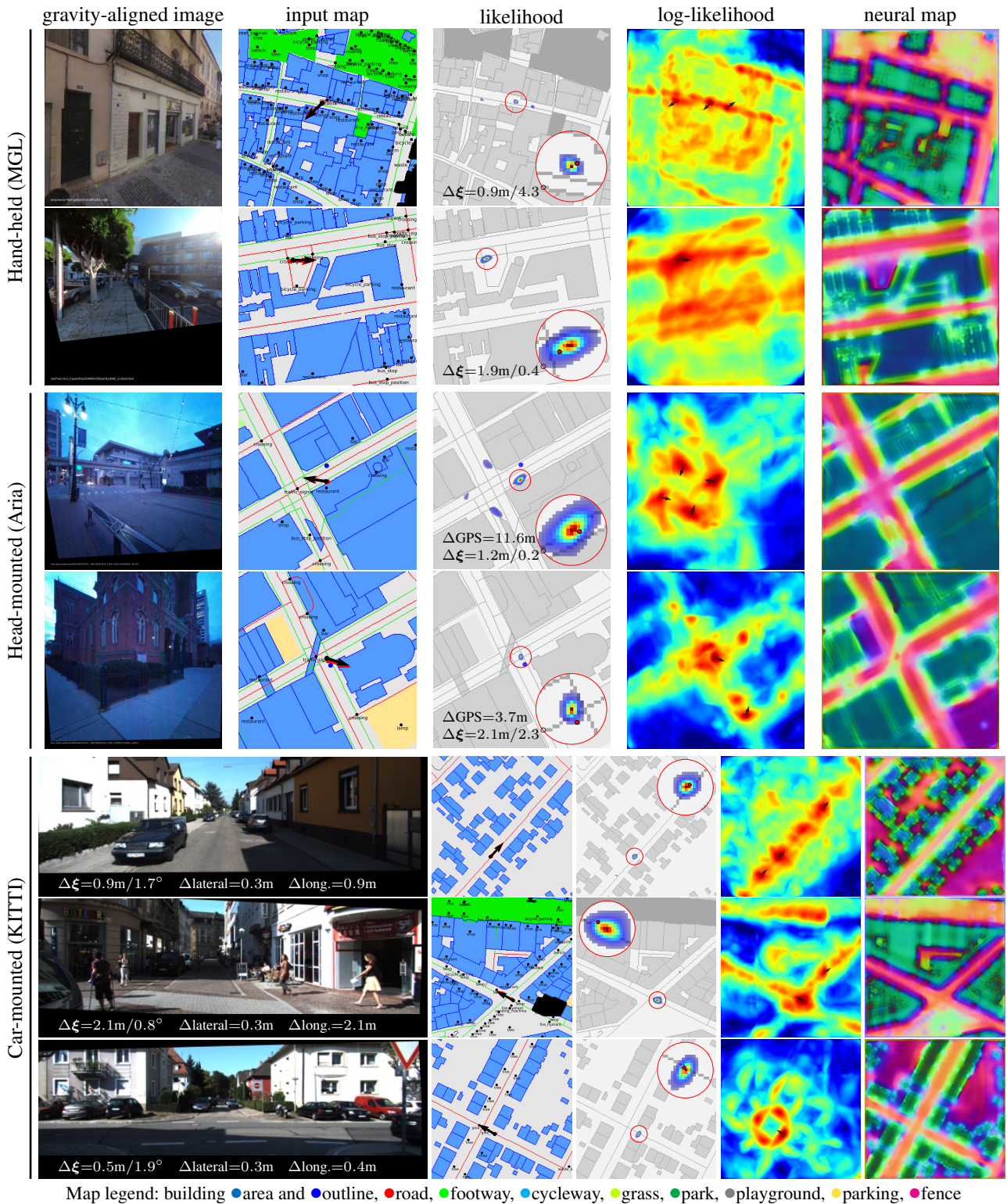

    \centering
    \input{figures/qualitative_supp_v2}
    \vspace{0.2cm}
    \caption{\textbf{Additional qualitative results for single-image localization.}
    We again show the pose predicted by OrienterNet (black~arrow \raisebox{-2pt}{\includegraphics[height=9pt]{figures/arrow-right.png}}) along with the ground truth pose (red arrow).
    For Aria data, we also show the noisy GPS measurement as a blue dot \cbullet{blue}.
    }%
    \vspace{0.5cm}
    \label{fig:qualitative-supp}%
\end{figure*}

\begin{figure*}[ht!]
    \centering
    \input{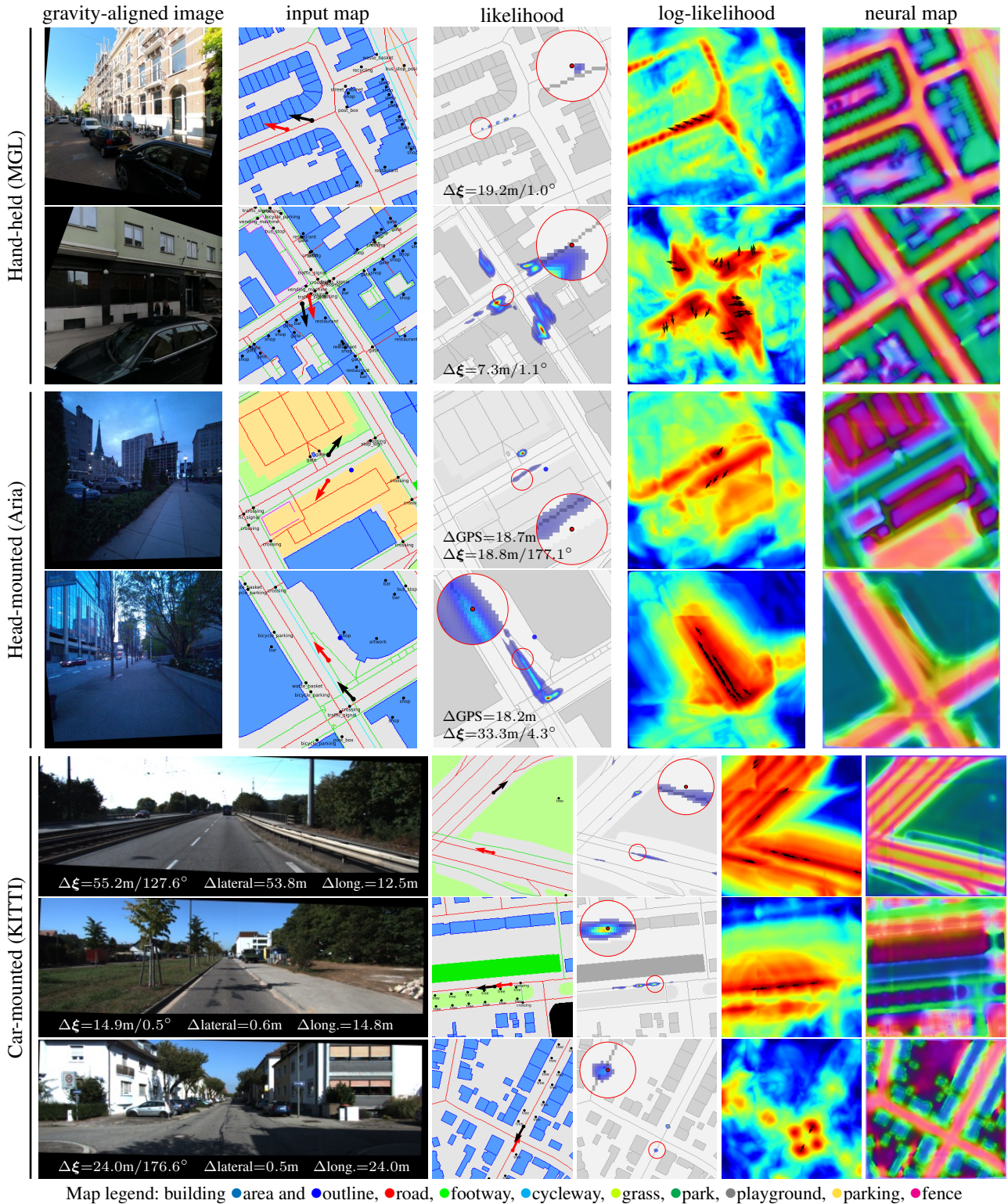}
    \vspace{0.2cm}
    \caption{\textbf{Failure cases of single-image localization.}
    Localizing a single image often fails when the environment lacks distinctive elements, when they do not appear in the map, or when such elements are repeated, making the pose ambiguous.
    Since OSM is crowd-sourced, the level of detail of the map is not consistent and widely varies.
    For example, trees are registered in some cities but not in others.
    }%
    \vspace{0.5cm}
    \label{fig:qualitative-supp-failures}%
\end{figure*}

\ifproceedings
\else
{\footnotesize%
\PAR{Acknowledgments:}
Thanks to Yann Noutary, Manuel López Antequera, Aleksander Colovic, and Arno Knapitsch for assisting with using OpenSfM and Mapillary data.
Thanks to Silvan Weder, Songyou Peng, and the anonymous reviewers for their thoughtful comments.
\par}
\fi

\fi 
\ifproceedings
    \ifwithsupp
        \newpage
        \appendix

\ifproceedings
\begin{center}
    \vspace{-1.4cm}
    {\Large \bf OrienterNet: Visual Localization in 2D Public Maps with Neural Matching\par}
    \vspace{0.5cm}
    {
        \large
        CVPR 2023 
    }
\end{center}
\fi

\section*{\supp}

\section{Additional results}
\label{sec:supp:results}

\PAR{Which map elements are most important?}
We study in \cref{fig:osm-ablation} the impact of each type of map element on the final accuracy by dropping them from the input map.
The classes with the largest impact are buildings and road, which are also the most common in areas covered by the training data.

\PAR{Impact of the field-of-view:}
We study the impact of the FoV on the accuracy by cropping the images in the horizontal direction to varying degrees.
Figure~\ref{fig:fov} shows the results on the MGL validation set.
Reducing the FoV decreases the accuracy proportionally -- a 50\% smaller FoV results in half of the original accuracy.

\PAR{Qualitative results:}
We show additional examples of single-image predictions in \cref{fig:qualitative-supp} and failure cases in \cref{fig:qualitative-supp-failures}.

\section{Data processing and distribution}

\subsection{OpenStreetMap}
\label{sec:supp:osm}
\PAR{Map classes:} 
OpenStreetMap~\cite{OpenStreetMap} exposes for each element a set of tags with standardized categories and labels according to a very rich hierarchy.
We group elements into a smaller set of classes that we list in \cref{tab:osm-classes}, resulting in 7 types of areas, 10 types of lines, and 33 types of points (nodes).
\Cref{fig:osm-distribution} shows the distribution of such elements for a small area.
\Cref{fig:osm-mapping} shows how some OSM tags are mapped to some semantic classes.

\PAR{Coordinate system:}
The coordinates of the map elements are given in WGS84 coordinates (longitude and latitude).
We convert them to a local scaled Mercator datum centered at the median camera pose of each area.
This yields topocentric coordinates that are aligned with the East and North axes.

\subsection{Mapillary Geo-Localization dataset}

\PAR{Curation process:}
We browsed the Mapillary platform and looked for sequences that were sufficiently recent and with the most accurate ground truth poses.
We selected sequences recorded after 2017 and with cameras known for resulting in good reconstructions.
These include the Xiaomi Yi Action 2K (fisheye) or GoPro Max, MADV QJXJ01FJ, or LG-R105 (spherical) cameras.
We selected 12 cities, listed in \cref{tab:mgl-loc}, that have a high density of such sequences.
\Cref{fig:mgl-loc} shows maps overlayed with the selected sequences.
Images of each location were split into disjoint training and validation sets, resulting in 826k training and 2k validation views.

\PAR{Preprocessing:}
We discard sequences with poor reconstruction statistics or high overlap with OSM building footprints.
We subsample the sequences such that frames are spaced by at least 4 meters.
We undistorted fisheye images into pinhole cameras.
We resampled each 360 panorama into 4 90\degree-FOV perspective views at equally-distributed yaw angles with a random offset constant per sequence.
We query OSM data for each city and create tiles of our raster representation at a resolution $\Delta{=}\SI{50}{\centi\meter}$.

\begin{figure}[t]
    \centering
    \includegraphics[width=\linewidth]{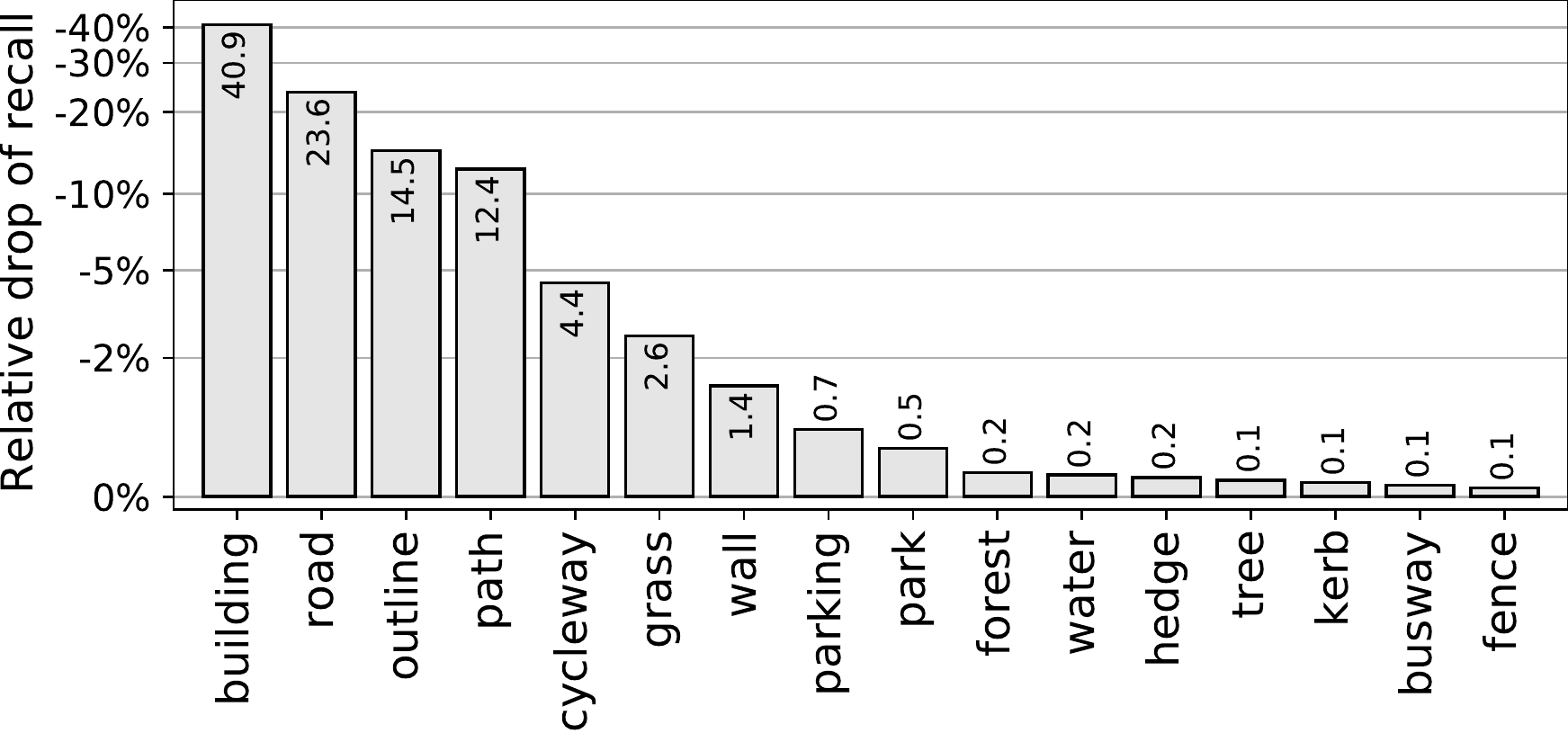}
    \caption{\textbf{Good semantics to localize.}
    Removing different elements from the map reveals how important they are for localization.
    Buildings, roads, footpaths, and cycleways are the most useful semantic classes, likely because they are also the most frequent.
    }%
\vspace{2mm}
    \label{fig:osm-ablation}%
\end{figure}

\begin{figure}[t]
    \centering
\begin{minipage}{0.49\linewidth}
\includegraphics[width=\linewidth]{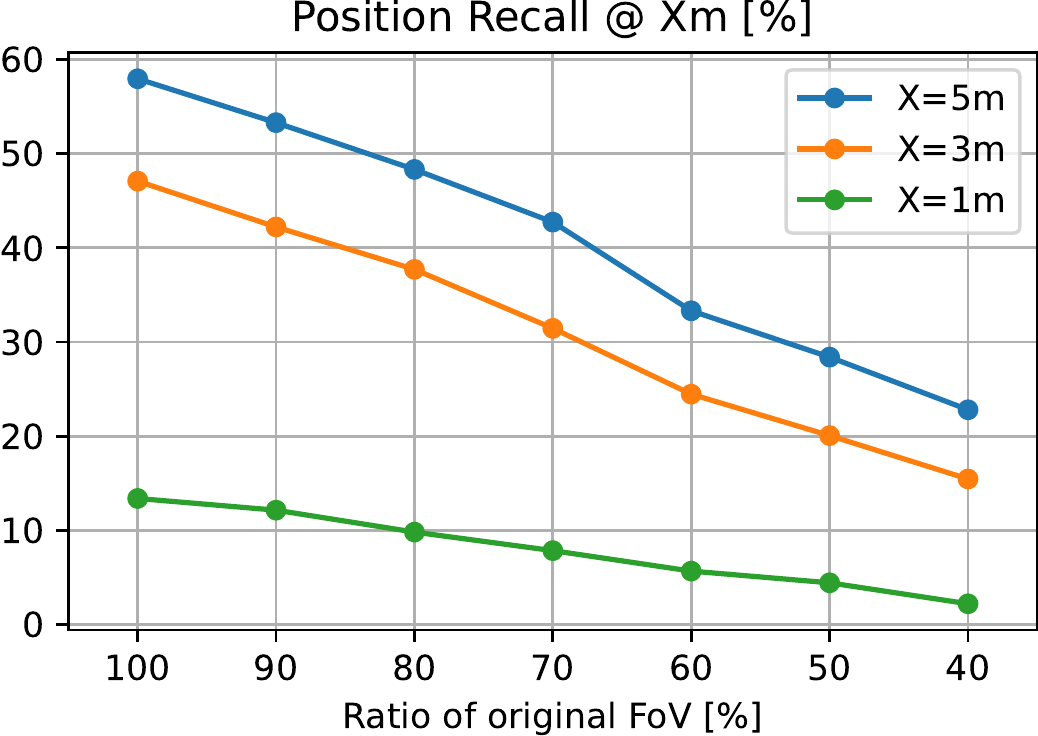}
\end{minipage}%
\hfill
\begin{minipage}{0.49\linewidth}
\includegraphics[width=\linewidth]{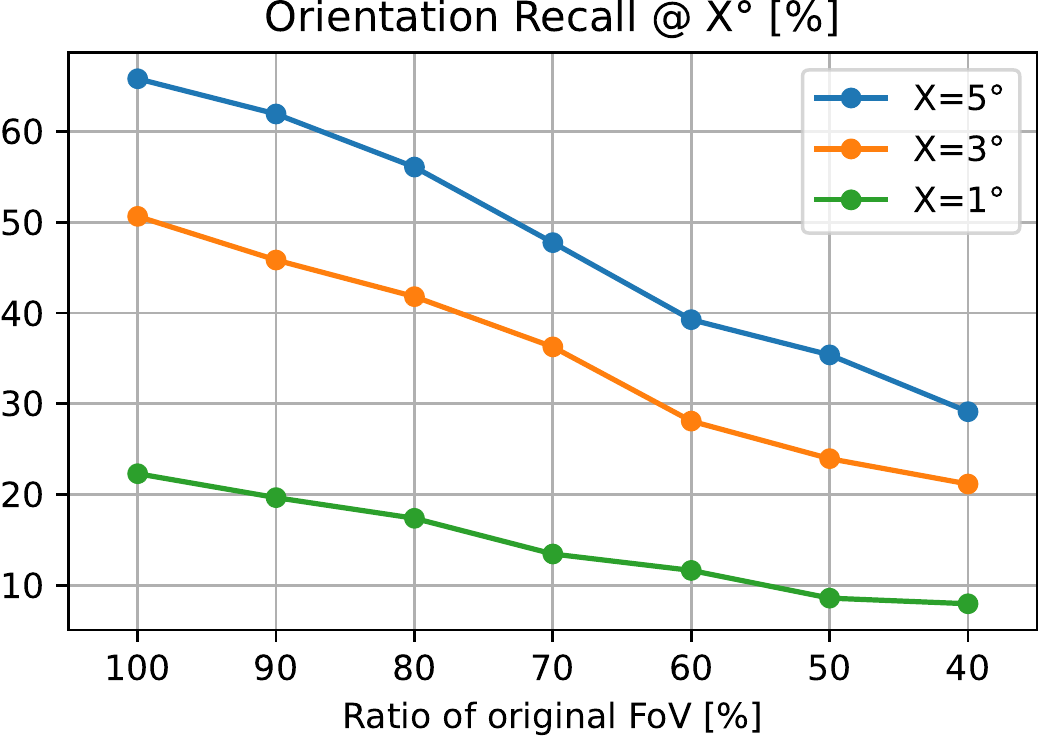}
\end{minipage}%
\vspace{1mm}
\caption{\textbf{Impact of the field of view} on the localization recall with the MGL validation set.
Decreasing the FoV directly impairs the accuracy as fewer map elements are visible in a single image.}
\vspace{2mm}
\label{fig:fov}
\end{figure}

\begin{table}[t]
\centering
\vspace{1mm}
\footnotesize{\begin{tabular}{ccc}
\toprule
type & classes\\
\midrule
areas & \makecell{parking spot/lot, building, grass, \\playground, park, forest, water}\\
\midrule
lines & \makecell{road, cycleway, pathway, busway, \\fence, wall, hedge, kerb, building outline, tree row}\\
\midrule
nodes & \makecell{parking entrance, street lamp, junction, traffic signal,\\stop sign, give way sign, bus stop, stop area, crossing, \\gate, bollard, gas station, bicycle parking, charging station, \\shop, restaurant, bar, vending machine, pharmacy,\\tree, stone, ATM, toilets, water fountain, bench,\\waste basket, post box, artwork, recycling station,\\clock, fire hydrant, pole, street cabinet}\\
\bottomrule
\end{tabular}}
\caption{\textbf{List of map classes derived from OpenStreeMap data} and included in the map rasters.
}%
\vspace{2mm}
\label{tab:osm-classes}
\end{table}

\begin{table}[t]
\centering
\footnotesize{\begin{tabular}{llcc}
\toprule
Country & City & \# sequences & \# images\\
\midrule
USA & San Francisco & 1013 & 207.6k\\
\midrule
Netherlands & Amsterdam & \0\057 & \072.9k\\
\midrule
Germany & Berlin & \0\059 & \054.6k\\
\midrule
Lithuania & Vilnius & \0381 & 111.5k\\
\midrule
Finland & Helsinki & \0\091 & \055.4k\\
\midrule
Italy & Milan & \0156 & \046.2k\\
\midrule
\multirow{6}{*}{France} & Paris & \0136 & \068.4k\\
& Montrouge & \0159 & \033.3k\\
& Le Mans & \0111 & \027.4k\\
& Nantes & \0171 & \062.5k\\
& Avignon & \0160 & \075.2k\\
& Toulouse & \0\086 & \039.6k\\
\bottomrule
\end{tabular}}
\vspace{2mm}
\caption{\textbf{Distribution of locations} from which we built the MGL dataset.
We selected cities that are well covered by both Mapillary and OpenStreetMap.
\vspace{0.3cm}
}%
\label{tab:mgl-loc}
\end{table}

\subsection{Aria datasets}
\PAR{Recording:}
We recorded data with Aria devices~\cite{aria} at 3 locations in Seattle (Downtown, Pike Place Market, Westlake) and at 2 locations in Detroit (Greektown, Grand Circus Park).
In each location, we recorded 3 to 5 sequences following the same trajectories, for a duration of 5 to 25 minutes varying by location.
Each device is equipped with a consumer-grade GPS sensor, IMUs, grayscale SLAM cameras, and a front-facing RGB camera, which we undistort to a pinhole model.

\PAR{Evaluation:}
We associate GPS signals captured at 1Hz with undistorted $640{\times}640$ RGB images keyframed at 3 meters.
This resulted in 2153 frames for Seattle and 2725 frames for Detroit.
For each evaluation example, the map tile is centered around the noisy GPS measurement.
Because of large differences in GPS accuracy due to urban canyons, we constrain the predictions within \SI{64}{\meter} of the measurement for Seattle and \SI{24}{\meter} for Detroit.

\PAR{Comparison to feature matching:}
Algorithms based on 3D SfM maps require mapping images, whose quality and density have a large impact on the localization accuracy.
Differently, OrienterNet can localize in areas not covered by such images as long as OSM data is available.
This makes any fair comparison difficult.

\PAR{Geo-alignment:}
Evaluating the localization within world-aligned maps requires the geo-alignement between each device pose and the global reference frame.
We first co-register all trajectories of each location by minimizing visual-inertial SLAM constraints, which yields consistent poses in a local reference frame that is gravity-aligned and at metric scale.
We then find the global 3-DoF rigid transformation by fusing all GPS signals with the predictions of OrienterNet for each image.
To do, we perform iterative truncated least-squares, annealing the outlier threshold from 5m to 1m.

We show visualizations of this alignment in \cref{fig:aria-alignment}.
While GPS signals are too noisy to reliably fit a transformation, OrienterNet provides complementary and accurate local constraints.
We visually check that the final alignment error is lower than 1m, which is sufficient for our evaluation.

\begin{figure}[t]
    \centering
    \input{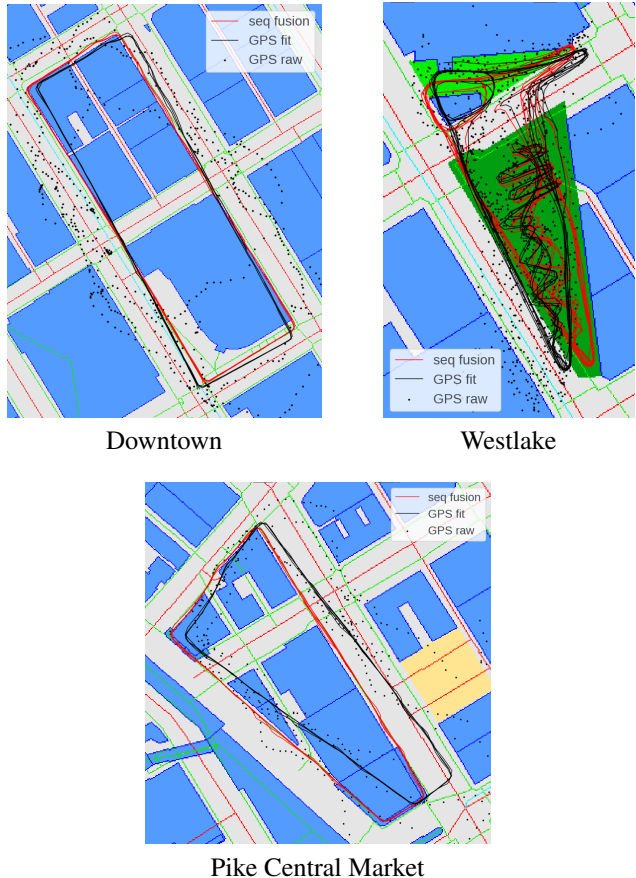}
    \vspace{3mm}
    \caption{\textbf{Pseudo-ground truth alignment of Aria sequences} for the 3 locations in Seattle.
    Fusing GPS signals and OrienterNet predictions across all images of all sequences is more robust than relying on GPS alone.
    }%
    \vspace{2mm}
    \label{fig:aria-alignment}%
\end{figure}

\section{Implementation details}
\PAR{Orienter-Net:}
To save GPU memory, we use only $K{=}64$ rotation bins at training time but increase it to $K{=}512$ at test time.
BEV and map features have $N{=}8$ channels.
To avoid overfitting, we found it critical to use replicate padding in the map-CNN $\Phi_\text{map}$ and to apply data augmentation to the raster map by randomly flipping and rotating it.
We also use replicate padding in the BEV-to-map matching operation to avoid biasing the predictions near the map boundaries.
The scale boundaries $\sigma_\text{min}$ and $\sigma_\text{max}$ are set to $2^1$ and $2^9$, respectively.
For the median focal length $f=256\text{px}$ of the MGL training set, this corresponds to a depth interval of $[\SI{0.5}{\meter},\SI{128}{\meter}]$.

Training images are resized to $512{\times}512$ pixels.
When evaluating on KITTI and Aria data, images are resized such that their focal length is $f=256\text{px}$.
We train with a batch size of 9 over 3 V100 GPUs with 16GB VRAM each.
We select the best model checkpoint with early stopping based on the validation loss.

\PAR{Retrieval baseline:}
The work of Samano \etal~\cite{samano2020you} infers a global descriptor for each map patch.
This is inefficient when considering densely sampled areas.
We can equivalently predict a dense feature map $\*F$ in one CNN forward pass, which is similar to the recent work of Xia \etal~\cite{xia2022visual} for satellite imagery.
We then correlate the global image descriptor with $\*F$ to obtain the pixelwise log-score $\*M$.
To predict a rotation, $\Phi_\text{map}$ computes 4 feature maps ${\*F_\text{N},\*F_\text{S},\*F_\text{E},\*F_\text{W}}$ for the 4 N-S-E-W directions, from which we can linearly interpolate map features for any number of rotation bins
This yields a 3D $W{\times}H{\times}K$ pose volume $\*M$ as for OrienterNet.
We found this approach much more efficient than re-computing map features for different map orientations.

\PAR{Refinement baseline:}
We follow the official implementation of Shi~\etal~\cite{shi2022beyond} with multi-level optimization at each iteration and warm restart.
We supervise the longitudinal, lateral, and angular offsets with an L2 loss at each iteration and scale.
The map data and branch are identical to OrienterNet.

\begin{figure*}[ht!]
    \centering
    \input{figures/mgl_locations}
    \vspace{0.3cm}
    \caption{\textbf{Selected sequences of our MGL dataset} across 12 cities.
    Screenshots taken from the Mapillary platform browser.
    }%
    \vspace{0.6cm}
    \label{fig:mgl-loc}%
\end{figure*}

\begin{figure*}[ht!]
    \vspace{2cm}
    \centering
    \includegraphics[width=\linewidth]{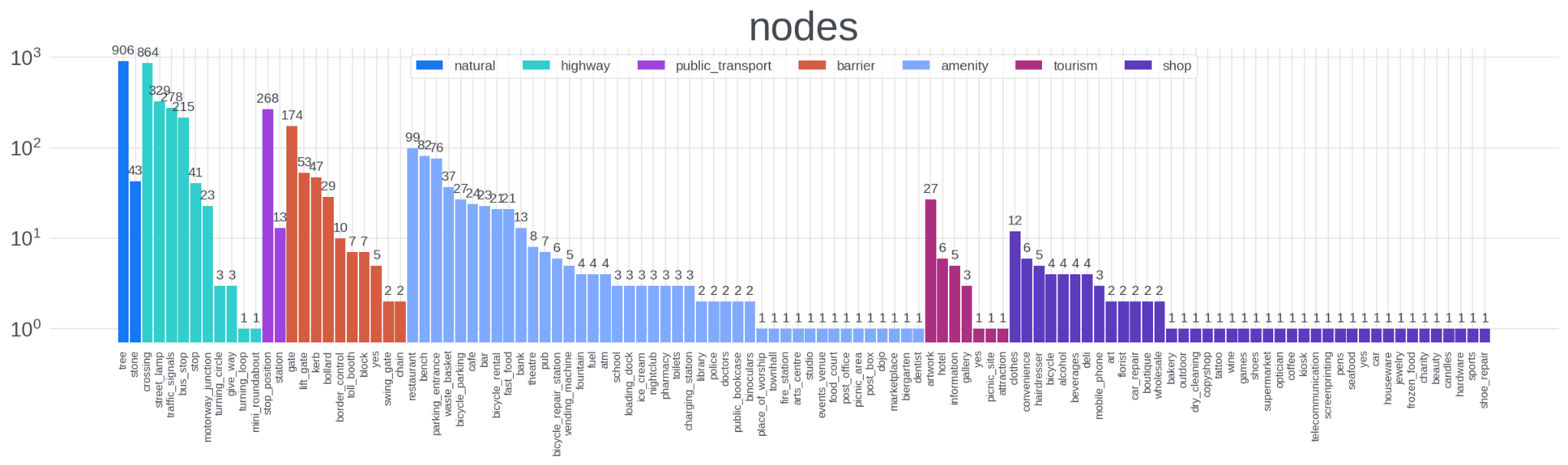}%
    \vspace{1cm}
    
    \includegraphics[width=\linewidth]{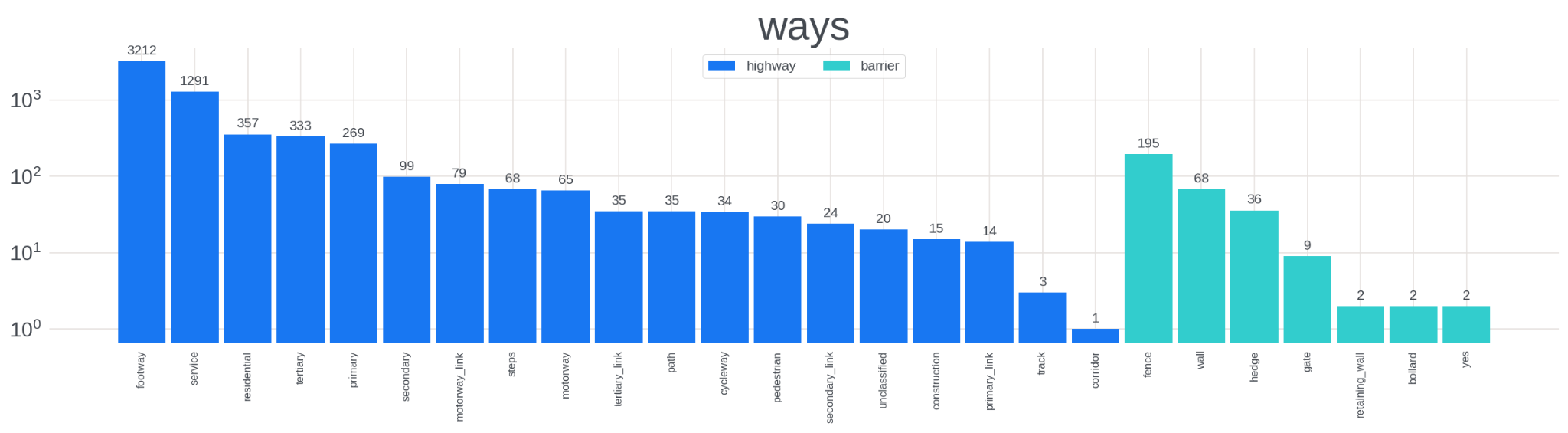}%
    \vspace{1cm}
    
    \includegraphics[width=\linewidth]{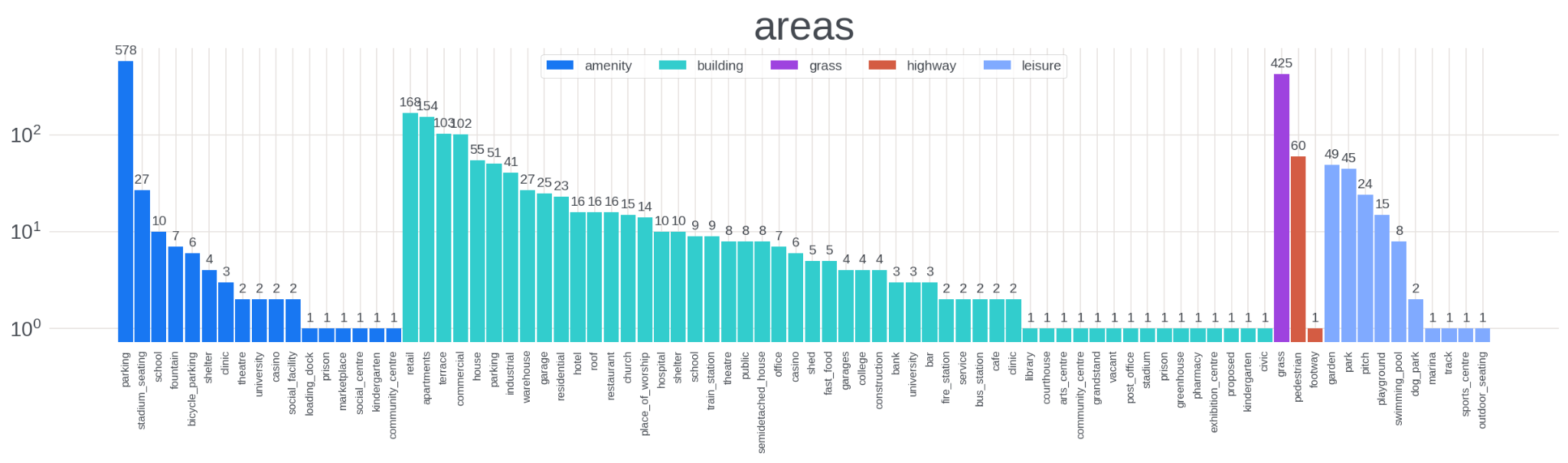}%
    \vspace{1cm}
    
    \caption{\textbf{Distribution of OSM elements} for the city of Detroit: points (top), lines (middle), polygons (bottom).
    We group them by label and order them by frequency.
    }%
    \vspace{1cm}
    \label{fig:osm-distribution}%
\end{figure*}

\begin{figure*}[ht!]
    \centering
    \includegraphics[width=0.48\linewidth]{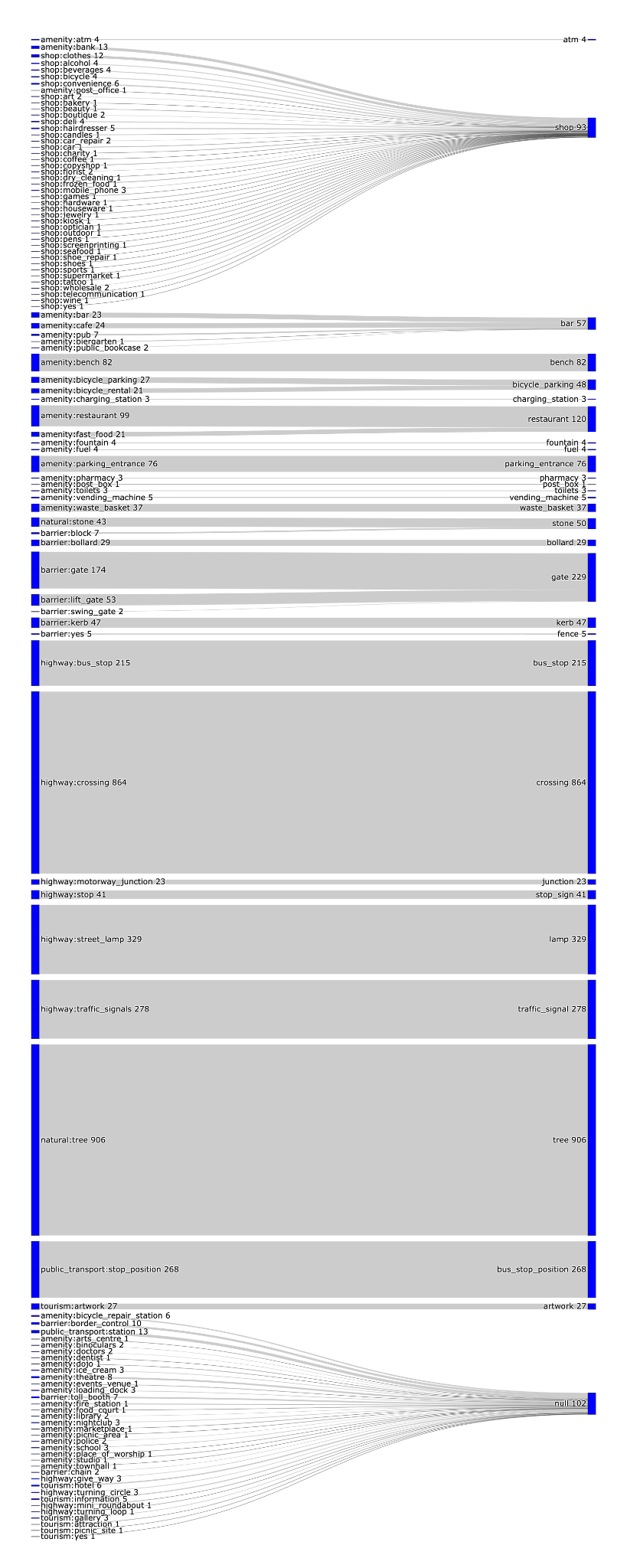}%
    
    \caption{\textbf{Mapping between OSM tags and raster classes}.
    We show a subset of the classes and the number of corresponding elements for the city of Detroit.
    }%
    \label{fig:osm-mapping}%
\end{figure*}
    \fi
\else
    
\fi

\ifproceedings
\ifsupponly
\else
\newpage
\clearpage
\fi
\else
\fi

{\small
\bibliographystyle{ieee_fullname}
\bibliography{mybib}

\begin{thebibliography}{10}\itemsep=-1pt

\bibitem{aria}
{Project Aria}.
\newblock \url{https://about.facebook.com/realitylabs/projectaria/}, 2022.

\bibitem{agarwal2011building}
Sameer Agarwal, Yasutaka Furukawa, Noah Snavely, Ian Simon, Brian Curless,
  Steven~M Seitz, and Richard Szeliski.
\newblock Building {Rome} in a day.
\newblock {\em Communications of the ACM}, 54(10):105--112, 2011.

\bibitem{antequera2020mapillary}
Manuel~L{\'o}pez Antequera, Pau Gargallo, Markus Hofinger, Samuel~Rota
  Bul{\`o}, Yubin Kuang, and Peter Kontschieder.
\newblock Mapillary planet-scale depth dataset.
\newblock In {\em ECCV}, 2020.

\bibitem{arandjelovic2016netvlad}
Relja Arandjelovic, Petr Gronat, Akihiko Torii, Tomas Pajdla, and Josef Sivic.
\newblock {NetVLAD}: {CNN} architecture for weakly supervised place
  recognition.
\newblock In {\em CVPR}, 2016.

\bibitem{armagan2017accurate}
Anil Armagan, Martin Hirzer, Peter~M Roth, and Vincent Lepetit.
\newblock Accurate camera registration in urban environments using high-level
  feature matching.
\newblock In {\em BMVC}, 2017.

\bibitem{armagan2017learning}
Anil Armagan, Martin Hirzer, Peter~M Roth, and Vincent Lepetit.
\newblock Learning to align semantic segmentation and {2.5D} maps for
  geolocalization.
\newblock In {\em CVPR}, 2017.

\bibitem{pmlr-v100-barnes20a}
Dan Barnes, Rob Weston, and Ingmar Posner.
\newblock Masking by moving: Learning distraction-free radar odometry from pose
  information.
\newblock In {\em CoRL}, 2020.

\bibitem{pmlr-v87-barsan18a}
Ioan~Andrei Barsan, Shenlong Wang, Andrei Pokrovsky, and Raquel Urtasun.
\newblock Learning to localize using a lidar intensity map.
\newblock In {\em CoRL}, 2018.

\bibitem{bay2006surf}
Herbert Bay, Tinne Tuytelaars, and Luc Van~Gool.
\newblock {SURF}: Speeded up robust features.
\newblock In {\em ECCV}, 2006.

\bibitem{Bujnak08CVPR}
Martin Bujnak, Zuzana Kukelova, and Tomas Pajdla.
\newblock A general solution to the p4p problem for camera with unknown focal
  length.
\newblock In {\em CVPR}, 2008.

\bibitem{burgard1996estimating}
Wolfram Burgard, Dieter Fox, Daniel Hennig, and Timo Schmidt.
\newblock Estimating the absolute position of a mobile robot using position
  probability grids.
\newblock In {\em AAAI}, 1996.

\bibitem{Camposeco_2019_CVPR}
Federico Camposeco, Andrea Cohen, Marc Pollefeys, and Torsten Sattler.
\newblock Hybrid scene compression for visual localization.
\newblock In {\em CVPR}, June 2019.

\bibitem{cao2014minimal}
Song Cao and Noah Snavely.
\newblock Minimal scene descriptions from structure from motion models.
\newblock In {\em CVPR}, 2014.

\bibitem{castaldo2015semantic}
Francesco Castaldo, Amir Zamir, Roland Angst, Francesco Palmieri, and Silvio
  Savarese.
\newblock Semantic cross-view matching.
\newblock In {\em ICCV Workshop on Vision from Satellite to Street}, 2015.

\bibitem{cham2010estimating}
Tat-Jen Cham, Arridhana Ciptadi, Wei-Chian Tan, Minh-Tri Pham, and Liang-Tien
  Chia.
\newblock Estimating camera pose from a single urban ground-view
  omnidirectional image and a {2D} building outline map.
\newblock In {\em CVPR}, 2010.

\bibitem{chu2014gps}
Hang Chu, Andrew Gallagher, and Tsuhan Chen.
\newblock {GPS} refinement and camera orientation estimation from a single
  image and a {2D} map.
\newblock In {\em CVPR Workshops}, 2014.

\bibitem{david2011Orientation}
Philip David and Sean Ho.
\newblock Orientation descriptors for localization in urban environments.
\newblock In {\em IROS}, 2011.

\bibitem{superpoint}
Daniel DeTone, Tomasz Malisiewicz, and Andrew Rabinovich.
\newblock {SuperPoint}: Self-supervised interest point detection and
  description.
\newblock In {\em CVPR Workshop on Deep Learning for Visual SLAM}, 2018.

\bibitem{dusmanu2019d2}
Mihai Dusmanu, Ignacio Rocco, Tomas Pajdla, Marc Pollefeys, Josef Sivic,
  Akihiko Torii, and Torsten Sattler.
\newblock {D2-Net}: A trainable {CNN} for joint detection and description of
  local features.
\newblock In {\em CVPR}, 2019.

\bibitem{Dusmanu2021Privacy}
Mihai Dusmanu, Johannes~L. Sch\"onberger, Sudipta~N. Sinha, and Marc Pollefeys.
\newblock {P}rivacy-{P}reserving {I}mage {F}eatures via {A}dversarial {A}ffine
  {S}ubspace {E}mbeddings.
\newblock In {\em CVPR}, 2021.

\bibitem{Floros2013OpenStreetSLAM}
Georgios Floros, Benito van~der Zander, and Bastian Leibe.
\newblock {OpenStreetSLAM}: Global vehicle localization using {OpenStreetMaps}.
\newblock In {\em ICRA}, 2013.

\bibitem{frahm2010building}
Jan-Michael Frahm, Pierre Fite-Georgel, David Gallup, Tim Johnson, Rahul
  Raguram, Changchang Wu, Yi-Hung Jen, Enrique Dunn, Brian Clipp, Svetlana
  Lazebnik, et~al.
\newblock Building {Rome} on a cloudless day.
\newblock In {\em ECCV}, 2010.

\bibitem{binarybow}
Dorian Galvez-López and Juan~D. Tardos.
\newblock Bags of binary words for fast place recognition in image sequences.
\newblock {\em IEEE Transactions on Robotics}, 28(5):1188--1197, 2012.

\bibitem{geiger2013vision}
Andreas Geiger, Philip Lenz, Christoph Stiller, and Raquel Urtasun.
\newblock Vision meets robotics: The {KITTI} dataset.
\newblock {\em The International Journal of Robotics Research},
  32(11):1231--1237, 2013.

\bibitem{guo2021coarse}
Chengcheng Guo, Minjie Lin, Heyang Guo, Pengpeng Liang, and Erkang Cheng.
\newblock Coarse-to-fine semantic localization with {HD} map for autonomous
  driving in structural scenes.
\newblock In {\em IROS}, 2021.

\bibitem{Haralick94IJCV}
R.M. Haralick, C.-N. Lee, K. Ottenberg, and M. N\"{o}lle.
\newblock Review and analysis of solutions of the three point perspective pose
  estimation problem.
\newblock {\em IJCV}, 13(3):331--356, 1994.

\bibitem{hong2019textplace}
Ziyang Hong, Yvan Petillot, David Lane, Yishu Miao, and Sen Wang.
\newblock {TextPlace}: Visual place recognition and topological localization
  through reading scene texts.
\newblock In {\em ICCV}, 2019.

\bibitem{howard2021lalaloc}
Henry Howard-Jenkins, Jose-Raul Ruiz-Sarmiento, and Victor~Adrian Prisacariu.
\newblock Lalaloc: Latent layout localisation in dynamic, unvisited
  environments.
\newblock In {\em CVPR}, 2021.

\bibitem{hoyer2021three}
Lukas Hoyer, Dengxin Dai, Yuhua Chen, Adrian Koring, Suman Saha, and Luc
  Van~Gool.
\newblock Three ways to improve semantic segmentation with self-supervised
  depth estimation.
\newblock In {\em CVPR}, 2021.

\bibitem{hu2018cvmnet}
Sixing Hu, Mengdan Feng, Rang M.~H. Nguyen, and Gim~Hee Lee.
\newblock {CVM-Net}: Cross-view matching network for image-based
  ground-to-aerial geo-localization.
\newblock In {\em CVPR}, 2018.

\bibitem{irschara2009structure}
Arnold Irschara, Christopher Zach, Jan-Michael Frahm, and Horst Bischof.
\newblock From structure-from-motion point clouds to fast location recognition.
\newblock In {\em CVPR}, 2009.

\bibitem{vlad}
Herv{\'e} J{\'e}gou, Matthijs Douze, Cordelia Schmid, and Patrick P{\'e}rez.
\newblock Aggregating local descriptors into a compact image representation.
\newblock In {\em CVPR}, 2010.

\bibitem{Kneip2011CVPR}
L Kneip, D Scaramuzza, and R Siegwart.
\newblock {A Novel Parametrization of the Perspective-Three-Point Problem for a
  Direct Computation of Absolute Camera Position and Orientation}.
\newblock In {\em CVPR}, 2011.

\bibitem{lao2022does}
Dong Lao, Alex Wong, and Stefano Soatto.
\newblock Does monocular depth estimation provide better pre-training than
  classification for semantic segmentation?
\newblock {\em arXiv:2203.13987}, 2022.

\bibitem{Larsson_2021_ICCV}
Viktor Larsson, Marc Pollefeys, and Magnus Oskarsson.
\newblock Orthographic-perspective epipolar geometry.
\newblock In {\em ICCV}, 2021.

\bibitem{lindenberger2021pixsfm}
Philipp Lindenberger, Paul-Edouard Sarlin, Viktor Larsson, and Marc Pollefeys.
\newblock {Pixel-Perfect Structure-from-Motion with Featuremetric Refinement}.
\newblock In {\em ICCV}, 2021.

\bibitem{lobben2007navigational}
Amy~K Lobben.
\newblock Navigational map reading: Predicting performance and identifying
  relative influence of map-related abilities.
\newblock {\em Annals of the association of American geographers},
  97(1):64--85, 2007.

\bibitem{lowe2004distinctive}
David~G Lowe.
\newblock Distinctive image features from scale-invariant keypoints.
\newblock {\em IJCV}, 60(2):91--110, 2004.

\bibitem{Lynen2020IJRR}
Simon Lynen, Bernhard Zeisl, Dror Aiger, Michael Bosse, Joel Hesch, Marc
  Pollefeys, Roland Siegwart, and Torsten Sattler.
\newblock {Large-scale, real-time visual-inertial localization revisited}.
\newblock {\em IJRR}, 39(9):1061--1084, 2020.

\bibitem{ma2019exploiting}
Wei-Chiu Ma, Ignacio Tartavull, Ioan~Andrei B{\^a}rsan, Shenlong Wang, Min Bai,
  Gellert Mattyus, Namdar Homayounfar, Shrinidhi~Kowshika Lakshmikanth, Andrei
  Pokrovsky, and Raquel Urtasun.
\newblock Exploiting sparse semantic {HD} maps for self-driving vehicle
  localization.
\newblock In {\em IROS}, 2019.

\bibitem{ma2017find}
Wei-Chiu Ma, Shenlong Wang, Marcus~A. Brubaker, Sanja Fidler, and Raquel
  Urtasun.
\newblock Find your way by observing the sun and other semantic cues.
\newblock In {\em ICRA}, 2017.

\bibitem{min2022laser}
Zhixiang Min, Naji Khosravan, Zachary Bessinger, Manjunath Narayana, Sing~Bing
  Kang, Enrique Dunn, and Ivaylo Boyadzhiev.
\newblock Laser: Latent space rendering for 2d visual localization.
\newblock In {\em CVPR}, 2022.

\bibitem{moulon2016openmvg}
Pierre Moulon, Pascal Monasse, Romuald Perrot, and Renaud Marlet.
\newblock Open{MVG}: Open multiple view geometry.
\newblock In {\em International Workshop on Reproducible Research in Pattern
  Recognition}, pages 60--74. Springer, 2016.

\bibitem{Ng_2022_CVPR}
Tony Ng, Hyo~Jin Kim, Vincent~T. Lee, Daniel DeTone, Tsun-Yi Yang, Tianwei
  Shen, Eddy Ilg, Vassileios Balntas, Krystian Mikolajczyk, and Chris Sweeney.
\newblock {NinjaDesc}: Content-concealing visual descriptors via adversarial
  learning.
\newblock In {\em CVPR}, 2022.

\bibitem{O'Keefe1978}
J. O'Keefe and L. Nadel.
\newblock {\em The hippocampus as a cognitive map}.
\newblock Clarendon Press, 1978.

\bibitem{OpenStreetMap}
{OpenStreetMap contributors}.
\newblock {Planet dump retrieved from https://planet.osm.org }.
\newblock \url{ https://www.openstreetmap.org }, 2017.

\bibitem{panphattarasap2018automated}
Pilailuck Panphattarasap and Andrew Calway.
\newblock Automated map reading: image based localisation in {2-D} maps using
  binary semantic descriptors.
\newblock In {\em IROS}, 2018.

\bibitem{pauls2020monocular}
Jan-Hendrik Pauls, Kürsat Petek, Fabian Poggenhans, and Christoph Stiller.
\newblock Monocular localization in {HD} maps by combining semantic
  segmentation and distance transform.
\newblock In {\em IROS}, 2020.

\bibitem{philion2020lift}
Jonah Philion and Sanja Fidler.
\newblock Lift, splat, shoot: Encoding images from arbitrary camera rigs by
  implicitly unprojecting to 3d.
\newblock In {\em ECCV}, 2020.

\bibitem{pittaluga2019revealing}
Francesco Pittaluga, Sanjeev~J Koppal, Sing~Bing Kang, and Sudipta~N Sinha.
\newblock Revealing scenes by inverting structure from motion reconstructions.
\newblock In {\em CVPR}, 2019.

\bibitem{revaud2019r2d2}
Jerome Revaud, Philippe Weinzaepfel, C{\'e}sar De~Souza, Noe Pion, Gabriela
  Csurka, Yohann Cabon, and Martin Humenberger.
\newblock {R2D2}: Repeatable and reliable detector and descriptor.
\newblock In {\em NeurIPS}, 2019.

\bibitem{roddick2020predicting}
Thomas Roddick and Roberto Cipolla.
\newblock Predicting semantic map representations from images using pyramid
  occupancy networks.
\newblock In {\em CVPR}, 2020.

\bibitem{rublee2011orb}
Ethan Rublee, Vincent Rabaud, Kurt Konolige, and Gary~R Bradski.
\newblock {ORB}: An efficient alternative to {SIFT} or {SURF}.
\newblock In {\em ICCV}, 2011.

\bibitem{ruchti2015Localization}
Philipp Ruchti, Bastian Steder, Michael Ruhnke, and Wolfram Burgard.
\newblock Localization on {OpenStreetMap} data using a 3d laser scanner.
\newblock In {\em ICRA}, 2015.

\bibitem{saha2022translating}
Avishkar Saha, Oscar Mendez, Chris Russell, and Richard Bowden.
\newblock Translating images into maps.
\newblock In {\em ICRA}, 2022.

\bibitem{samano2020you}
Noe Samano, Mengjie Zhou, and Andrew Calway.
\newblock You are here: Geolocation by embedding maps and images.
\newblock In {\em ECCV}, 2020.

\bibitem{sarlin2019coarse}
Paul-Edouard Sarlin, Cesar Cadena, Roland Siegwart, and Marcin Dymczyk.
\newblock From coarse to fine: Robust hierarchical localization at large scale.
\newblock In {\em CVPR}, 2019.

\bibitem{sarlin2020superglue}
Paul-Edouard Sarlin, Daniel DeTone, Tomasz Malisiewicz, and Andrew Rabinovich.
\newblock {SuperGlue}: Learning feature matching with graph neural networks.
\newblock In {\em CVPR}, 2020.

\bibitem{sarlin21pixloc}
Paul-Edouard Sarlin, Ajaykumar Unagar, Måns Larsson, Hugo Germain, Carl Toft,
  Viktor Larsson, Marc Pollefeys, Vincent Lepetit, Lars Hammarstrand, Fredrik
  Kahl, and Torsten Sattler.
\newblock {Back to the Feature}: Learning robust camera localization from
  pixels to pose.
\newblock In {\em CVPR}, 2021.

\bibitem{sattler2012improving}
Torsten Sattler, Bastian Leibe, and Leif Kobbelt.
\newblock Improving image-based localization by active correspondence search.
\newblock In {\em ECCV}, 2012.

\bibitem{schoenberger2016sfm}
Johannes~Lutz Sch\"{o}nberger and Jan-Michael Frahm.
\newblock Structure-from-motion revisited.
\newblock In {\em CVPR}, 2016.

\bibitem{shi2022beyond}
Yujiao Shi and Hongdong Li.
\newblock Beyond cross-view image retrieval: Highly accurate vehicle
  localization using satellite image.
\newblock In {\em CVPR}, 2022.

\bibitem{NEURIPS2019_ba2f0015}
Yujiao Shi, Liu Liu, Xin Yu, and Hongdong Li.
\newblock Spatial-aware feature aggregation for image based cross-view
  geo-localization.
\newblock In {\em NeurIPS}, 2019.

\bibitem{shi2020looking}
Yujiao Shi, Xin Yu, Dylan Campbell, and Hongdong Li.
\newblock Where am {I} looking at? joint location and orientation estimation by
  cross-view matching.
\newblock In {\em CVPR}, 2020.

\bibitem{shi2020optimal}
Yujiao Shi, Xin Yu, Liu Liu, Tong Zhang, and Hongdong Li.
\newblock Optimal feature transport for cross-view image geo-localization.
\newblock In {\em AAAI}, 2020.

\bibitem{SimmonsK95}
Reid~G. Simmons and Sven Koenig.
\newblock Probabilistic robot navigation in partially observable environments.
\newblock In {\em IJCAI}, 1995.

\bibitem{snavely2006photo}
Noah Snavely, Steven~M Seitz, and Richard Szeliski.
\newblock {Photo Tourism}: exploring photo collections in {3D}.
\newblock In {\em SIGGRAPH}, 2006.

\bibitem{SpecialeCVPR2019}
Pablo Speciale, Johannes~L. Sch{\"{o}}nberger, Sing~Bing Kang, Sudipta~N.
  Sinha, and Marc Pollefeys.
\newblock Privacy preserving image-based localization.
\newblock In {\em CVPR}, 2019.

\bibitem{specialeiccv2019}
Pablo Speciale, Johannes~L. Sch{\"{o}}nberger, Sudipta~N. Sinha, and Marc
  Pollefeys.
\newblock Privacy preserving image queries for camera localization.
\newblock In {\em ICCV}, 2019.

\bibitem{svarm2017city}
Linus Sv{\"a}rm, Olof Enqvist, Fredrik Kahl, and Magnus Oskarsson.
\newblock City-scale localization for cameras with known vertical direction.
\newblock {\em TPAMI}, 2017.

\bibitem{tang2021ijrr}
Tim~Y. Tang, Daniele~De Martini, Shangzhe Wu, and Paul Newman.
\newblock Self-supervised learning for using overhead imagery as maps in
  outdoor range sensor localization.
\newblock {\em IJRR}, 40(12-14):1488--1509, 2021.

\bibitem{torii201524}
Akihiko Torii, Relja Arandjelovic, Josef Sivic, Masatoshi Okutomi, and Tomas
  Pajdla.
\newblock 24/7 place recognition by view synthesis.
\newblock In {\em CVPR}, 2015.

\bibitem{tyszkiewicz2020disk}
Micha{\l}~J Tyszkiewicz, Pascal Fua, and Eduard Trulls.
\newblock {DISK}: Learning local features with policy gradient.
\newblock In {\em NeurIPS}, 2020.

\bibitem{vojir2020Efficient}
Tomas Vojir, Ignas Budvytis, and Roberto Cipolla.
\newblock Efficient large-scale semantic visual localization in {2D} maps.
\newblock In {\em ACCV}, 2020.

\bibitem{vysotska2017improving}
Olga Vysotska and Cyrill Stachniss.
\newblock Improving {SLAM} by exploiting building information from publicly
  available maps and localization priors.
\newblock {\em PFG--Journal of Photogrammetry, Remote Sensing and
  Geoinformation Science}, 85(1):53--65, 2017.

\bibitem{weng2021semantic}
Li Weng, Val{\'e}rie Gouet-Brunet, and Bahman Soheilian.
\newblock Semantic signatures for large-scale visual localization.
\newblock {\em Multimedia Tools and Applications}, 2021.

\bibitem{xia2022visual}
Zimin Xia, Olaf Booij, Marco Manfredi, and Julian~FP Kooij.
\newblock Visual cross-view metric localization with dense uncertainty
  estimates.
\newblock In {\em ECCV}, 2022.

\bibitem{yan2019global}
Fan Yan, Olga Vysotska, and Cyrill Stachniss.
\newblock Global localization on {OpenStreetMap} using 4-bit semantic
  descriptors.
\newblock In {\em ECMR}, 2019.

\bibitem{Zeisl_2015_ICCV}
Bernhard Zeisl, Torsten Sattler, and Marc Pollefeys.
\newblock Camera pose voting for large-scale image-based localization.
\newblock In {\em ICCV}, 2015.

\bibitem{zhou2021efficient}
Mengjie Zhou, Xieyuanli Chen, Noe Samano, Cyrill Stachniss, and Andrew Calway.
\newblock Efficient localisation using images and {OpenStreetMaps}.
\newblock In {\em IROS}, 2021.

\bibitem{zhou2022geometry}
Qunjie Zhou, Sergio Agostinho, Aljosa Osep, and Laura Leal-Taixe.
\newblock Is geometry enough for matching in visual localization?
\newblock In {\em ECCV}, 2022.

\bibitem{zhu2021vigor}
Sijie Zhu, Taojiannan Yang, and Chen Chen.
\newblock {VIGOR}: Cross-view image geo-localization beyond one-to-one
  retrieval.
\newblock In {\em CVPR}, 2021.

\end{thebibliography}
}

\end{document}